\documentclass[10pt,twocolumn,letterpaper]{article}

\usepackage[pagenumbers]{iccv} %

\usepackage{booktabs}
\usepackage{multirow}
\usepackage{pifont}
\usepackage{placeins}
\usepackage{siunitx}
\usepackage{xcolor,colortbl}

\newcommand{\cmark}{\ding{51}}%
\newcommand{\xmark}{\ding{55}}%

\definecolor{bg-blue}{HTML}{dbd6ef}

\definecolor{iccvblue}{rgb}{0.21,0.49,0.74}
\usepackage[pagebackref,breaklinks,colorlinks,allcolors=iccvblue]{hyperref}

\def\paperID{8565} %
\def\confName{ICCV}
\def\confYear{2025}

\usepackage{xspace}

\newcommand{\Model}{4DS\xspace}

\title{Scaling 4D Representations}

\author{Jo\~ao Carreira$^{\dag}$ \and Dilara Gokay$^{\dag}$ \and Michael King$^{\dag}$ \and Chuhan Zhang$^{\dag}$ \and Ignacio Rocco$^{\dag}$ \and Aravindh Mahendran$^{\dag}$ \and Thomas Albert Keck$^{\dag}$ \and Joseph Heyward$^{\dag}$ \and Skanda Koppula$^{\dag}$ \and Etienne Pot$^{\dag}$ \and Goker Erdogan$^{\dag}$ \and Yana Hasson$^{\dag}$ \and Yi Yang$^{\dag}$ \and Klaus Greff$^{\dag}$ \and Guillaume Le Moing$^{\dag}$ \and Sjoerd van Steenkiste$^{\ddag}$ \and Daniel Zoran$^{\dag}$ \and Drew A. Hudson$^{\dag}$ \and Pedro Vélez$^{\dag}$\and Luisa Polanía$^{\dag}$ \and Luke Friedman$^{\dag}$ \and Chris Duvarney$^{\dag}$ \and Ross Goroshin$^{\dag \star}$ \and Kelsey Allen$^{\dag}$ \and Jacob Walker$^{\dag}$ \and Rishabh Kabra$^{\dag}$ \and Eric Aboussouan$^{\dag}$ \and Jennifer Sun$^{\dag}$ \and Thomas Kipf$^{\dag}$ \and Carl Doersch$^{\dag}$ \and Viorica P\u atr\u aucean$^{\dag}$ \and Dima Damen$^{\dag \Delta}$ \and Pauline Luc$^{\dag}$ \and Mehdi S.\ M.\ Sajjadi$^{\dag}\;\,$ and $\;\,$Andrew Zisserman$^{\dag\Diamond}$ \\[3mm]
\small{
$^{\dag}$Google DeepMind \;\;\;
$^{\ddag}$Google Research \;\;\;
${^\star}$Mila/Université de Montréal \;\;\;
${^\Delta}$University of Bristol \;\;\; $^{\Diamond}$University of Oxford}\vspace{-4mm}\\
}

\begin{document}
\maketitle
\begin{abstract}
Scaling has not yet been convincingly demonstrated for pure self-supervised learning from video. However, prior work has focused evaluations on semantic-related tasks -- action  classification, ImageNet classification, etc. In this paper we focus on evaluating self-supervised learning on non-semantic vision tasks that are more spatial (3D) and temporal (+1D = 4D), such as camera pose estimation, point and object tracking, and depth estimation. We show that by learning from very large video datasets, masked auto-encoding (MAE) with transformer video models actually scales, consistently improving performance on these 4D tasks, as model size increases from 20M all the way to the largest by far reported self-supervised video model -- 22B parameters. Rigorous apples-to-apples comparison with many recent image and video models demonstrates the benefits of  scaling 4D representations. Pretrained models are available at~\href{https://github.com/google-deepmind/representations4d}{https://github.com/google-deepmind/representations4d}.
\end{abstract}
    
\vspace{-4mm}
\section{Introduction}
\vspace{-1mm}
\label{sec:intro}

Self-supervised learning of representations from video took the back seat in recent years to language-supervised approaches deriving from CLIP. Is the perceived superiority of language supervision because it is a better way to learn? Or may it have to do with dominant evaluations being also language-based – such as action classification and localization? Do we not yet have sufficiently powerful self-supervised learning techniques in our repertoire? Or just need to throw more hardware and data at them?

In this paper, we attempt to answer these questions by revisiting self-supervised learning, focusing the evaluation on four semantics-free tasks: depth estimation, point and object tracking, camera pose estimation. These are tasks that require strong spatial and temporal abilities -- they are less about naming things, more about recovering the 3D geometry over time of (4D) scenes. Making an analogy with human vision, we are interested in inspecting {\em dorsal} as opposed to {\em ventral} stream skills. Most prior work has focused on classification downstream tasks such as Something-something v2~\cite{goyal2017something} and Kinetics~\cite{kay2017kineticshumanactionvideo}, so we include results on those as well.

\begin{figure}
    \centering
    \vspace{1mm}
    \includegraphics[width=\linewidth]{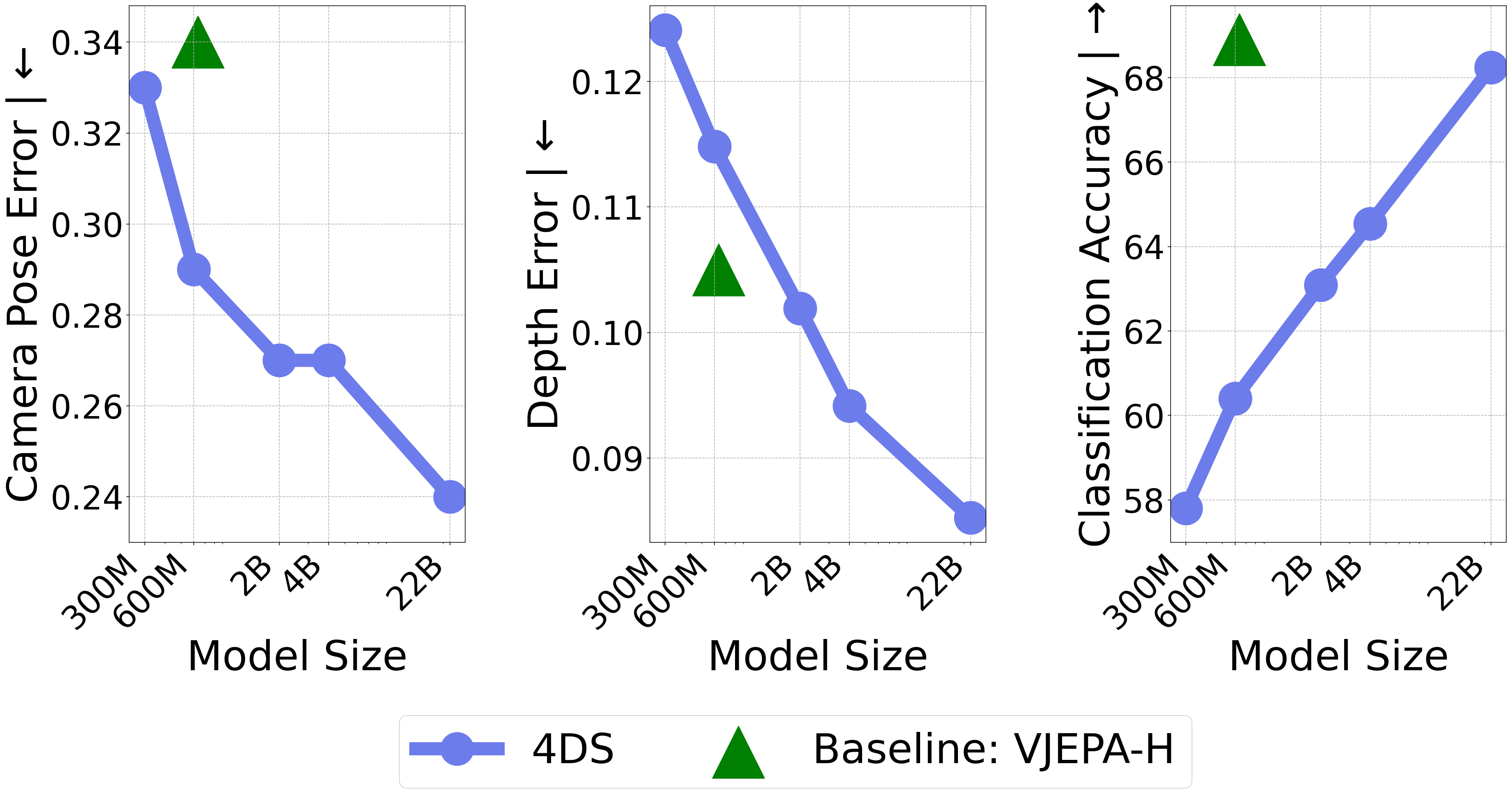}
    \caption{Performance keeps improving on all downstream video tasks as we scale self-supervised video models to up to 22B parameters (shown: frozen feature evaluation). We find that simple scaling of ViTs with masked auto-encoding already delivers for non-semantic video tasks requiring accurate 4D representations such as RE10K camera pose estimation~\cite{realestate10k} (left, lower is better) and ScanNet depth estimation~\cite{dai2017scannet} (middle, lower is better). No prior self-supervised video models have ventured beyond 1B parameters, but they have focused on semantic tasks such as SSv2 action classification~\cite{goyal2017something} (right, higher is better).\vspace{-3mm}}
    \label{fig:scaling_plot}
\end{figure}

We bring together some of the strongest models available -- both image (DinoV2~\cite{oquab2023dinov2}, SigLIP~\cite{chen2023pali}) and video models (VideoPrism~\cite{zhao2024videoprism}, VideoMAE-v1/v2~\cite{tong2022videomae,wang2023videomae}, V-JEPA~\cite{bardes2023vjepa}), pretrained with and without language supervision -- and evaluate them, carefully ensuring these are apples-to-apples comparisons. We employ for all tasks and models the same attention-based readouts, on top of either frozen backbones or using full finetuning. 

Our conclusions in all settings are:

\begin{enumerate}
\item Image models are not competitive. Even though our readouts are spatio-temporal, the lack of temporal representation in the backbone reveals itself as an unsurmountable problem.
\item VideoMAE and V-JEPA both perform well. More recent pretraining schemes may result in strong performance on popular classification tasks, such as Kinetics~\cite{kay2017kineticshumanactionvideo}, but underperform on the 4D tasks.
\item Language supervision by itself seems inferior to video self-supervision.
\end{enumerate}

We opted to dive into MAE and scale up transformer models using it, from 20M all the way to the by far largest self-supervised video model reported so far -- 22B parameters, far higher than existing public self-supervised video models (no text) that top out at 1B parameters. We train on a large collection of 170M 30-second videos using a custom bare-bones MAE approach we call SimpleMAE.
We observe consistent overall improvement on all tasks as we increase model size, far beyond the 600M parameter setting past which performance saturation has been observed for semantics-centric evaluations~\cite{wang2023videomae}.
It has been more widely a common belief\footnote{Based on personal communication with many elements in the community -- unfortunately negative results do not tend to be formally published.} in the community that MAE has mediocre scaling properties, to which our work brings nuance.

\vspace{2mm}
\noindent \textbf{Contributions.} Our contributions are three-fold: a) a re-evaluation of the state-of-the-art from the perspective of 4D scene representation quality;  b) SimpleMAE: a frugal approach to video MAE that removes complexities such as target normalization, tube masking and custom decoders which produced c) three new MAE-VIT models beyond the largest existing 1B ones, having 2B, 4B and 22B parameters, a model family which we term \Model. Inference code and representative checkpoints are released online at \href{https://github.com/google-deepmind/representations4d}{https://github.com/google-deepmind/representations4d}, and also include a 90M model distilled from the 4B one.

\vspace{2mm}
\noindent \textbf{Paper structure.} We go over related work in the next section. We then explain the paper methodology: SimpleMAE, baseline models, tasks, readouts and evaluation  in~\cref{sec:methodology}. \cref{sec:results} contains all the results and learnings before the paper concludes in~\cref{sec:conclusions}.

\section{Related work}
\label{sec:related_work}

\noindent\textbf{Origins.}
Video is a privileged medium for unsupervised learning -- a high-bandwidth connection to the world, providing multiple viewpoints of the same objects through time, across lighting changes, deformation and so on. Much early work in self-supervised learning in computer vision focused on video, for example by leveraging temporal continuity signals~\cite{foldiak1991learning,wiskott2002slow,mobahi2009deep}, later boosted by object tracking information~\cite{Wang_2015_ICCV}, motion segmentation~\cite{pathak2017learning}, camera pose between frames~\cite{Agrawal_2015_ICCV}, audio~\cite{arandjelovic17look,arandjelovic2018objects,Senocak_2018_CVPR,owens2018audio,korbar2018cooperative,alwassel2019self,mandela2020datatrans,morgado20avid}, future prediction~\cite{ranzato2014video,vondrick2015anticipating,goroshin2015learning}, contrastive learning under augmentation~\cite{hjelm2020representation,han2020memory,han2020self,qian2021spatiotemporal} and regression-based approaches~\cite{Recasens_2021_ICCV,feichtenhofer2021large}.
Context prediction gave strong results as supervision for  text~\cite{collobert2008unified}, and soon proved effective for visual representation learning~\cite{doersch2015unsupervised,pathak2016context,henaff2020data}.
With the rise of Vision Transformers~\cite{dosovitskiy2020image}, MAE became a particularly efficient formulation of context prediction~\cite{he2022masked}, since masked patches can be skipped by the encoder.

\vspace{2mm}
\noindent\textbf{Scaling.} Video versions of masked modeling achieved strong results~\cite{tong2022videomae, bardes2023vjepa} but the excitement was not long-lasting as attempts to scale models like MAE have yet to succeed (there are no published models beyond 1B parameters). In parallel, CLIP~\cite{radford2021learningtransferablevisualmodels} took over, enabling learning from billions of labels from the web, and this translated well also to mainstream video tasks~\cite{wang2022internvideo,zhao2024videoprism} such as classification~\cite{kay2017kineticshumanactionvideo} and action localization~\cite{caba2015activitynet} (which are defined using words as labels).
Models evaluated on the basis of their features (e.g.\ representations) have been convincingly scaled using language supervision~\cite{dehghani2023scaling} for images. Attempts to scale from just pixels resulted in more modest improvements~\cite{pmlr-v119-chen20s,el2024scalable}. In the video space, GPT-like~\cite{radford2018improving} vision and language video models have demonstrated clear benefit from scaling~\cite{alayrac2022flamingo}. Pure video self-supervised models have not yet been scaled anywhere near to comparable scales, with some of the biggest models having around 1B parameters~\cite{wang2023videomae}, or 3B for  auto-regressive models~\cite{Bai_2024_CVPR} -- this is something now explored with this paper. Related to our approach, VideoMAE v2~\cite{wang2023videomae} scales a video MAE model, but focuses solely on action recognition benchmarks. They observe only small performance improvements when scaling from ViT-H to ViT-g, and therefore do not scale models beyond the billion parameter regime.

\vspace{2mm}
\noindent\textbf{\textit{Dorsal} streams and evaluation.} The dorsal stream arises in the two-streams hypothesis~\cite{mishkin1983object,Goodale92,felleman1991distributed}, that the human visual cortex contains two pathways: the ventral stream (which performs object recognition) and the dorsal stream (which is sensitive to transformations and motion). It has been the (indirect) inspiration for several of the early architectures for video representation~\cite{simonyan2014,feichtenhofer2019slowfast,Feichtenhofer_nips_2016}. Several recent papers studied representations on 3D tasks~\cite{El_Banani_2024_CVPR,zhan2024generalprotocolprobelarge}, with results pointing to shortcomings in models solely supervised with language but considered only images. Benchmarks such as Physion~\cite{bear2022physionevaluatingphysicalprediction} and the Perception Test~\cite{patraucean2023perception} focus on videos and measure more closely the 4D capabilities we are interested in (we adopt one of the tasks from the latter).

\section{Methodology}
\label{sec:methodology}

 \begin{table}
 \begin{small}
\centering
\caption{\label{tab:vit_configs} \textbf{\Model model collection.} The  7 space-time VIT model configurations trained in the paper, sorted by number of parameters (\#p). The last column lists the total number of parameters (\#tp) including decoding parameters -- note there are very few of them in our design. The encoders match VIT image models~\cite{dosovitskiy2020image,dehghani2023scaling} except for the first layer which is a 3D instead of 2D convolution (the corresponding input space-time patch size is 2x16x16 in all cases). The j configuration is new.} 
\begin{tabular}{crrrrrr}
\toprule
{} &  {\small Width} &  {\small Depth} &  {\small MLP}  &  {\small \#heads}  & {\small \#p (M)} & {\small \#tp (M)} \\
\midrule
S &       384 &      12 &       1536 &      6  & 23 & 24 \\
B &       768 &      12 &       3072 &     12  & 88 & 91 \\
L &      1024 &      24 &       4096 &     16  & 306 & 310 \\
H &      1280 &      32 &       5120 &     16  & 634 & 639\\
G &      1664 &      48 &       8192 &     16  & 1,848 & 1,854 \\
e &      1792 &      56 &      15360 &     16  & 3,811 & 3,817 \\
j &      4096 &      64 &      32768 &     32  & 21,495 & 21,509 \\
\bottomrule
\end{tabular}
\end{small}
\end{table}

Our goal is to re-evaluate a large set of strong models apples-to-apples on 4D representation tasks that focus on geometry and motion. To achieve this we follow a representation learning framework where pretrained models get appended with attention-based readouts on top and are then separately finetuned for each of the various downstream datasets / tasks, with and without freezing of the backbone. In our experiments detailed in~\cref{sec:results} we found MAE to do well and to be efficiently trainable so we scaled ViT models in that framework -- details on scaling are given in the next subsection, followed by the overall evaluation protocol including readout architectures and finally a list of baselines.

\subsection{Learning from video with SimpleMAE}
\label{sec:simplemae}

MAE works by masking out a subset of input patches to an encoder, then having a decoder produce back the masked patches~\cite{pathak2016context}. Previous image MAE implementations additionally employed a) target normalization~\cite{he2022masked}, where target patches are individually normalized; and b) separate lightweight decoders with custom hyperparameters~\cite{he2022masked}. Video MAE added to the mix c) tube masking~\cite{tong2022videomae}, where temporal tubes of patches are masked all at once. We do none of these in our training approach -- SimpleMAE --  isolating instead the impact of the scale of the models and data on performance, without technical bells and whistles. Using SimpleMAE we train a family of 7 models, that we name \Model, spanning 20M to 22B parameters, listed in~\cref{tab:vit_configs}.

\vspace{2mm}
\noindent \textbf{No special masking.} We mask in all cases randomly 95\% of 2x16x16 space-time patches (no tricks like tube masking or any special masking pattern), feeding the remaining to the transformer model. 

\begin{figure}
    \centering
    \includegraphics[width=\columnwidth]{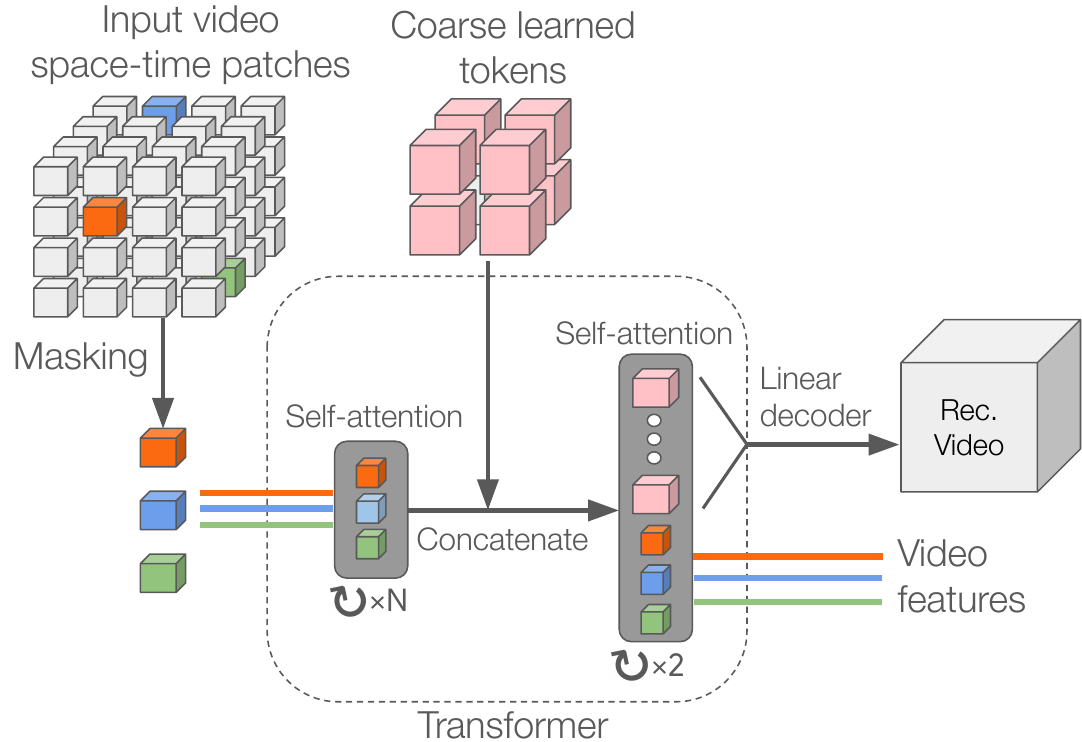}
    \caption{\textbf{SimpleMAE framework.} Videos are encoded into space-time tokens via one non-overlapping convolution typical of ViTs, then 95\% are randomly dropped. A transformer model then processes the remaining tokens with self-attention layers; first alone, then jointly with a learned grid of latent tokens on the last few self-attention layers (last two for the 22B model). Each of these learned tokens is decoded back to a patch using a linear layer. We do not use a custom separate decoder, nor target normalization, nor tube masking. For downstream evaluations, video features from any layer can be input to additional trainable light-weight readouts.\vspace{-3mm}}
    \label{fig:model}
\end{figure}

\vspace{2mm}
\noindent \textbf{No custom decoder.} We use the last few self-attention blocks of the encoder to do decoding work.  The process is illustrated in~\cref{fig:model}: we concatenate a learned grid of latent tokens to the unmasked set of tokens before one of the last self-attention blocks. We then decode these extra tokens back to pixel space linearly and patchwise. In downstream tasks, the latent tokens are discarded; features can be extracted from any layer. For models up to 4B parameters we use a decoding grid of tokens the same shape as the input video grid ($8 \times 14 \times 14 = 1568$, 8 for time, 14x14 for space, for 16 frames at 224x224 resolution). For 22B we use a more efficient, coarser grid ($4 \times 8 \times 8 = 256$ tokens for 16 frames at 256x256 resolution). We reconstruct all space-time patches, including unmasked ones.

\vspace{2mm}
\noindent \textbf{No target normalization.} We use an L2 loss between RGB values without any special tricks (e.g.\ no contrast normalization~\cite{he2022masked,tong2022videomae}).

\begin{figure*}
    \centering
    \begin{subfigure}[t]{0.19\linewidth}
        \includegraphics[width=\linewidth]{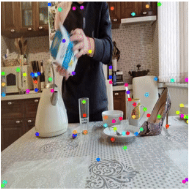}
        \captionsetup{width=0.96\linewidth}
        \caption{Point tracking on the Perception Test dataset~\cite{patraucean2023perception}.}
    \end{subfigure}
    \begin{subfigure}[t]{0.192\linewidth}
        \includegraphics[width=\linewidth]{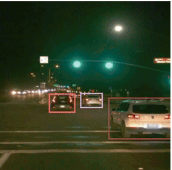}
        \captionsetup{width=0.96\linewidth}
        \caption{Object tracking on the Waymo Open dataset~\cite{9156973}.}
    \end{subfigure}
    \begin{subfigure}[t]{0.178\linewidth}
        \includegraphics[width=\linewidth]{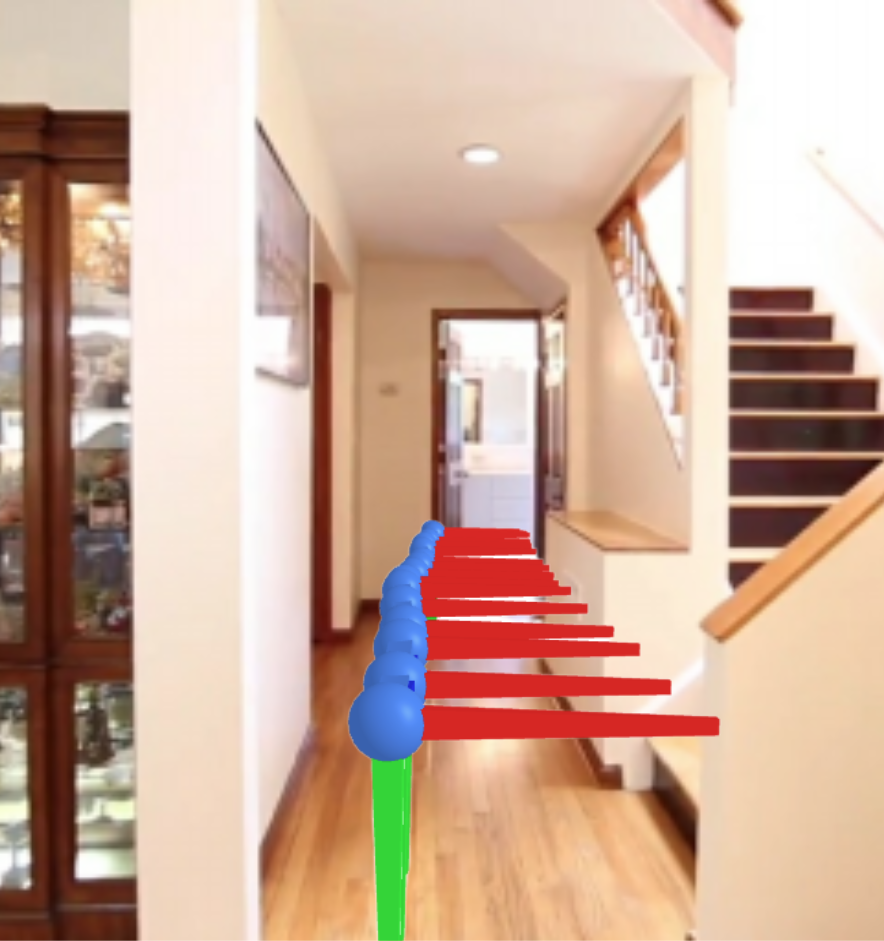}
        \captionsetup{width=0.96\linewidth}
        \caption{Camera pose estimation on the RealEstate10k dataset~\cite{realestate10k}.}
    \end{subfigure}
    \begin{subfigure}[t]{0.19\linewidth}
        \includegraphics[width=\linewidth]{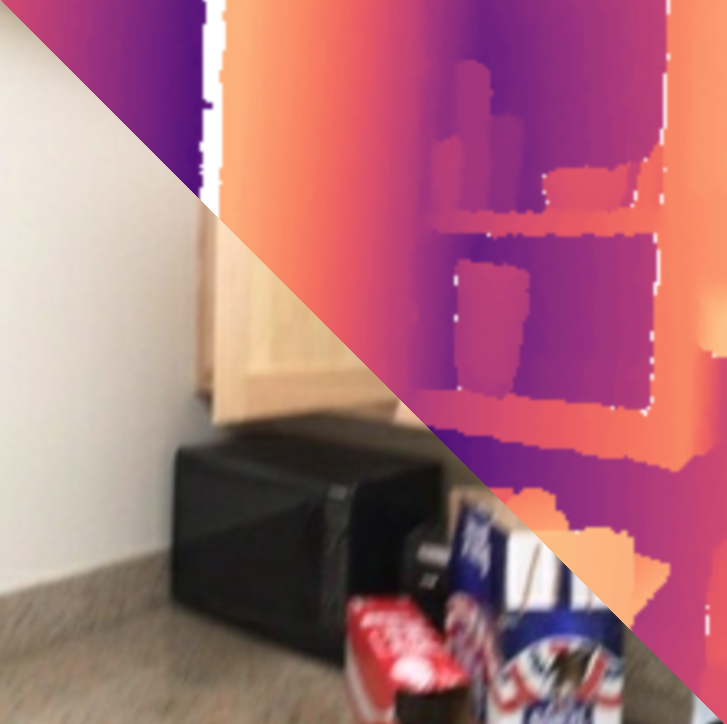}
        \captionsetup{width=0.96\linewidth}
        \caption{Depth estimation on the ScanNet dataset~\cite{dai2017scannet}.
        }
    \end{subfigure}
    \begin{subfigure}[t]{0.18\linewidth}
        \includegraphics[width=\linewidth]{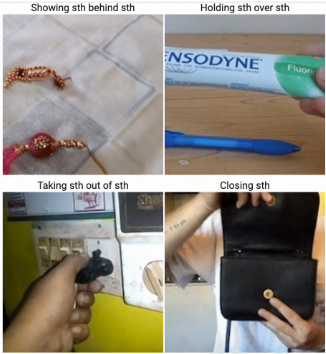}
        \captionsetup{width=0.96\linewidth}
        \caption{Action classification on the SSv2~\cite{goyal2017something} and Kinetics~\cite{kay2017kineticshumanactionvideo,smaira2020short} datasets.}
    \end{subfigure}
    \caption{\label{fig:tasks} \textbf{Evaluation downstream tasks}. Individual frames and annotations from the five tasks considered in this paper. The majority of previous work has focused on classification, whereas in this paper we do a deep dive into the other four tasks that are less semantic and more about spatial and temporal (4D) scene perception.\vspace{-2mm}}
\end{figure*}

\subsection{Training setup}

We pretrain \Model  models using a dataset of 170 million web videos, each 900 frames long on average, at original resolution and frame rate. We sample batches of 16 frame clips from these videos, using temporal stride 2 and randomly cropping down to 224x224 (models up to 4B parameters) or 256x256 resolution (22B parameter models) after resizing the smallest side to 1.15x those resolutions. Models up to 4B parameters are trained on 1 billion clips, and the 22B models on 2 billion clips, all 16 frames long and sampled randomly from the long 170M videos -- smaller models see significantly more data relative to their number of parameters, which benefits them.

We train each model using 256 TPUs-v6, with \emph{bfloat16} precision except for the loss and softmax computation which is upcast to \emph{float32}; weights are stored in \emph{float32}. The standard AdamW optimizer~\cite{loshchilov2019decoupledweightdecayregularization} is used. We employ simple FSDP-like sharding~\cite{fsdp} of model parameters and optimizer state in order to abide by per-device memory limits (32GB). Additional details including all hyperparameters for training are provided in the supplementary material.

\subsection{Evaluation protocol}

We followed an evaluation protocol designed to fairly compare arbitrary backbones over multiple tasks in the same setting. We tried both end-to-end finetuning and frozen evaluation, which puts more emphasis on the quality of the pretraining and is much cheaper computationally. For each task we add a trainable attention-based readout module on top of a pretrained video encoder then measure the resulting performance on the validation split. Compared to linear readouts, attention-based ones offer much better performance on top of frozen features (sometimes close to full finetuning) while not increasing parameter count significantly and have become preferred recently~\cite{bardes2023vjepa} The input to the model is a 16-frame clip  which is preprocessed (e.g.\ resampled) to match each model's preferred input format, then the output features are of shape $T\times K\times C$ where $T$ is a time dimension, $K$ is the number of spatial tokens per time step and $C$ is the number of channels for that model. We resample $T$ to be 16 as required by some of the task readout modules requirements. A $T \times K \times C$-shaped learnable positional embedding is added to the  features in order to not disadvantage image encoders. Readouts are trained using 1.28M examples and AdamW~\cite{loshchilov2019decoupledweightdecayregularization}. No masking, noising, etc.\ is applied to the videos on any of the tasks -- additional details are given in the supplementary material.

\vspace{2mm}
\noindent \textbf{Tasks.} We focus on tasks requiring strong 4D representations. They are shown in~\cref{fig:tasks} and capture different aspects of spatial and temporal scene understanding: (a) point tracking on Perception Test~\cite{patraucean2023perception} measures fine-grained, temporally extended motion perception; (b) object tracking from bounding boxes in the Waymo Open Dataset~\cite{9156973} emphasizes object-level motion. We also include two tasks that connect with 3D perception: (c) 6D camera pose estimation on the Real Estate 10K dataset (RE10k)~\cite{realestate10k,zhou2018stereo} and (d)  monocular depth estimation in ScanNet~\cite{dai2017scannet}. To connect with prior work, we also include (e) action classification in Kinetics-700-2020~\cite{smaira2020short} and SSv2~\cite{goyal2017something} -- the latter is well known to require temporal sensitivity (e.g.\ shuffling frames leads to very poor performance).  All of the tasks are defined in the literature except for the RE10k task for which we introduce a custom metric as well -- we give full details about all the tasks including this new one in the supp. material.

\vspace{2mm}
\noindent \textbf{Attention-based readouts.} There are several common strategies for evaluation of representation learning methods on downstream tasks, particularly with frozen backbones, ranging from non-parametric (nearest-neighbor)~\cite{oquab2023dinov2}, linear or using cross-attention layers~\cite{bardes2023vjepa, zhao2024videoprism,tong2022videomae}. In this work we use cross-attention-based readout heads for all tasks: for SSv2, Kinetics, and RE10k a single learned query is used as in~\cite{bardes2023vjepa}, employing an additional step of Procrustes-based projection~\cite{bregier2021deep} back into $SO(3)$ for RE10k. For ScanNet we use Fourier-based spatial cross-attention queries, one for each 2x8x8 patch. For the tracking tasks we employ one query per object track for Waymo Open, and one query for each 2-frame temporal chunk of a point track for Perception Test (so 8 queries per track for 16 frame clips). Full details of readout head architectures and hyper-parameters are provided in supp. material.

\subsection{Baselines}
\label{sec:baselines}
For baselines we include some of the strongest and largest image and video models available -- we detail them below, see~\cref{tab:baselines} for a quick reference.

\begin{table*}[t]
\centering
\caption{Comparison of existing (first two blocks of rows) and new models (\Model, last set of rows) when learning an attention readout on top of a \textbf{frozen backbone}. \textbf{Bold} indicates best performance, \underline{underlined} indicates second-best. All results are obtained using the same evaluation protocol, with exact same inputs, same trainable readouts, and using exact same (limited) number of training examples -- see text for details. ScanNet values are multiplied by 10 for better readability. First 4 tasks are the 4D perception tasks we focus on (and where the larger 4DS models do best). We include on the right SSv2 and K700-2020, which are semantic, classification tasks, to connect with prior work that focused on those. 4DS-B-dist-e is a B model distilled from 4DS-e.}
\setlength{\tabcolsep}{3pt}
\small{
\begin{tabular}{l@{\hskip -0.05in}c@{\hskip 0.05in}c@{\hskip 0.05in}c@{\hskip 0in}|cccc|cc}
\toprule
\textbf{Model} & Size (M) & \multirow{2}{*}{\parbox{1.6cm}{\centering Image (I)/ Video (V)}} & \multirow{2}{*}{\parbox{2cm}{\centering Language pre-training}} & \textbf{RE10k} & \multirow{2}{*}{\parbox{1.6cm}{\centering \textbf{Percep. test ($\uparrow$)}}} & \textbf{ScanNet} & \multirow{2}{*}{\parbox{1.6cm}{\centering \textbf{Waymo Open ($\uparrow$)}}} & \textbf{SSv2} & \textbf{K700-2020} \\ 
& & & &($\downarrow$)& &($\downarrow$)& &($\uparrow$) &($\uparrow$)\\
\midrule
SigLIP \cite{chen2023pali}      & 	1,705    & I & \cmark & 2.71             & 40.2             & 1.47             & 45.5 & 50.5  & 66.4 \\ 
DinoV2-L \cite{oquab2023dinov2}     & 303  & I & \xmark & 2.00             & 36.6             & 1.02 & 51.7 & 52.2  &  63.5 \\
DinoV2-g \cite{oquab2023dinov2}    & 1,135 & I & \xmark & 1.89 & 39.9            & \underline{0.91} & 50.6 & 54.8 & 65.4 \\
\midrule
VideoPrism-B \cite{zhao2024videoprism} & 114  & V & \cmark & 0.93             & 71.9 & 1.20             & 66.6 & 60.7   & 64.3 \\ 
VideoPrism-g \cite{zhao2024videoprism} & 1,113  & V & \cmark & 0.90             & 75.0    & 0.96    & 68.5 & 65.4 & \textbf{70.3} \\
V-JEPA-L \cite{bardes2023vjepa}    & 307  & V & \xmark & 0.39             & 75.6             & 1.14             & 73.3 & 66.0  & 55.2 \\ 
V-JEPA-H \cite{bardes2023vjepa}    & 635  & V & \xmark & 0.34             & 78.0             & 1.02             & 74.9 & \textbf{68.9} & 57.0 \\ 
VideoMAE-B \cite{tong2022videomae}   & 87 & V & \xmark & 0.37             & 79.2 & 1.51             & 73.1 & 52.3  & 38.9 \\ 
VideoMAE-L \cite{tong2022videomae}   & 305  & V & \xmark & 0.28 & 78.3 & 1.09            & 74.9 & 62.7 & 52.5 \\
VideoMAE-H  \cite{tong2022videomae}  & 633  & V & \xmark & 0.30    & 77.1             & 1.03 & 74.6 & 64.2 & 54.5 \\ 
VideoMAEv2-g \cite{wang2023videomae}~~~ & 1,013    & V & \cmark & 0.51       & 73.7             & 1.08 & 72.6 & 65.6 & \underline{69.7} \\ 
\midrule
\midrule
\Model-S & 24  & V & \xmark & 0.73    & 75.9 & 2.05             & 69.6 & 39.9 & 28.1 \\
\Model-B & 91  & V & \xmark & 0.51    & 78.9 & 1.65             & 72.7 & 49.6 & 35.7 \\
\Model-B-dist-e & 91 & V & \xmark & 0.34    & 81.4 & 1.21         & 76.3  & 60.3 & 46.4 \\
\Model-L & 310  & V & \xmark & 0.33    & 81.5 & 1.23             & 75.9 & 57.6 & 45.2 \\
\Model-H & 639  & V & \xmark & 0.29    & 81.8 & 1.14            & 76.1 & 60.0 & 47.5 \\
\Model-G & 1,848  & V & \xmark & \underline{0.27}    & \underline{82.7} & 1.01             & 77.4 & 62.5  & 52.1 \\
\Model-e & 3,811  & V & \xmark & \underline{0.27}   & 82.4 & 0.95 & \underline{78.0} & 64.3  & 54.4 \\
\Model-j & 21,495 & V & \xmark & \textbf{0.24}    & \textbf{83.4}    & \textbf{0.84}    &  \textbf{78.3} & \underline{68.2} & 58.0 \\
\bottomrule
\end{tabular}}
\label{tab:baselines}
\end{table*}

\begin{table*}[t]
\centering
\caption{Same as~\cref{tab:baselines} but with \textbf{short finetuning} for 20k steps (nothing is frozen). The trends are consistent with frozen evaluation, but most results are better in absolute terms.}
\setlength{\tabcolsep}{3pt}
\small{
\begin{tabular}{l@{\hskip -0.05in}c@{\hskip 0.05in}c@{\hskip 0.05in}c@{\hskip 0in}|cccc|cc}
\toprule
\textbf{Model} & Size (M) & \multirow{2}{*}{\parbox{1.6cm}{\centering Image (I)/ Video (V)}} & \multirow{2}{*}{\parbox{2cm}{\centering Language pre-training}} & \textbf{RE10k} & \multirow{2}{*}{\parbox{1.6cm}{\centering \textbf{Percep. test ($\uparrow$)}}} & \textbf{ScanNet} & \multirow{2}{*}{\parbox{1.6cm}{\centering \textbf{Waymo Open ($\uparrow$)}}} & \textbf{SSv2} & \textbf{K700-2020} \\ 
& & & &($\downarrow$)& &($\downarrow$)& &($\uparrow$) &($\uparrow$)\\
\midrule
SigLIP \cite{chen2023pali}      & 	1,705    & I & \cmark & 2.84             & 32.9             & 1.33             & 44.9 & 57.4   & 67.7 \\ 
DinoV2-L \cite{oquab2023dinov2}    & 303  & I & \xmark & 2.00            & 48.3            & 0.85 & 50.8 & 61.0 & 62.4 \\
DinoV2-g \cite{oquab2023dinov2}    & 1,135 & I & \xmark & 2.09 & 40.5           & \underline{0.72} & 50.2 & 64.2 & 64.8 \\
\midrule
VideoPrism-B \cite{zhao2024videoprism} & 114  & V & \cmark & 0.81            & 72.8 & 1.07             & 66.6 & 63.6  & 59.6 \\ 
VideoPrism-g \cite{zhao2024videoprism} & 1,113  & V & \cmark & 0.81            & 73.7   & 0.87   & 67.1 & 68.6 & \textbf{71.2} \\
V-JEPA-L \cite{bardes2023vjepa}    & 307  & V & \xmark & 0.35             & 76.0             & 0.94             & 74.1 & \underline{70.4} & 56.8 \\ 
V-JEPA-H \cite{bardes2023vjepa}    & 635  & V & \xmark & 0.29            &  71.5           & 0.85            & 74.6 & \textbf{72.9} & 57.6 \\ 
VideoMAE-B \cite{tong2022videomae}   & 87 & V & \xmark & 0.36             & 79.7 & 1.50             & 72.0 & 54.8 & 36.4 \\ 
VideoMAE-L \cite{tong2022videomae}  & 305  & V & \xmark & \underline{0.28} & 79.1 & 0.98             & 74.0 & 65.6  & 50.3 \\
VideoMAE-H \cite{tong2022videomae}   & 633  & V & \xmark & 0.32    & 77.3             & 0.93 & 73.3 & 66.5 & 53.0 \\ 
VideoMAEv2-g \cite{wang2023videomae}~~~ & 1,013    & V & \cmark & 0.42            & 76.7           & 0.91 & 71.1 & 70.1 & \underline{70.1} \\ 
\midrule
\midrule
\Model-S & 24  & V & \xmark & 0.72    & 75.2 & 2.05            & 69.1 & 40.3 & 25.4 \\
\Model-B & 91  & V & \xmark & 0.48    & 79.5 & 1.49             & 71.8 & 52.4 & 34.9 \\
\Model-L & 310  & V & \xmark & 0.37    & 81.1 & 1.07             & 74.8 & 61.8 & 45.2 \\
\Model-H & 639  & V & \xmark & 0.36    & 81.9 & 0.96           & 75.7 & 64.9 & 48.6 \\
\Model-G & 1,848  & V & \xmark & 0.29   & \underline{83.0} & 0.81             & 77.2 & 67.4 & 53.0 \\
\Model-e & 3,811  & V & \xmark & \underline{0.28}   & 82.9 & 0.73 & \underline{77.5} & 68.9 & 52.7 \\
\Model-j & 21,495 & V & \xmark & \textbf{0.26}   & \textbf{84.2}    &  \textbf{0.62}    &  \textbf{78.0} & \textbf{72.9} & 57.3 \\
\bottomrule
\end{tabular}}
\label{tab:baselines_ft}
\end{table*}

\vspace{2mm}
\noindent \textbf{Image models.} We evaluate \textit{SigLIP-2B}, part of PaLI-3~\cite{chen2023pali} and a follow-up of SigLIP~\cite{zhai2023sigmoid}. It is a large 2B vision transformer model trained from image-text pairs via a binary classification formulation. We also evaluate the extremely powerful and popular image model \textit{Dino-V2}~\cite{oquab2023dinov2} in two sizes: VIT-L and VIT-g, with $303$M and $1.1$B parameters, respectively. Dino-V2 is trained very differently and quite complementary to SigLIP-2B, using a self-distillation loss from images only (no text). SigLIP-2B is one of the strongest descendants of CLIP, Dino-V2 of SimCLR~\cite{chen2020simple}. Both models are trained on large web-scale image datasets.

Note again that we add learnable temporal positional embeddings to the  features produced by image models so they are not in principle disadvantaged when task readouts are trained on top in the frozen evaluation setting.

\vspace{2mm}
\noindent \textbf{Video models.} We evaluated VideoPrism~\cite{zhao2024videoprism}, a state-of-the-art video model combining two training objectives -- a) contrastive learning using video and language and b) video-only masked modelling. After the contrastive phase, a student model is trained to predict embeddings from masked frames similar to the predictions of the contrastively trained teacher from unmasked frames. VideoPrism is based on the ViViT architecture~\cite{vivit}, with space and time factorization and we tested two versions: B and g with 100M and 1B parameters, respectively.

We also evaluated two types of popular video-only self-supervised models: VideoMAE~\cite{tong2022videomae, wang2023videomae} and V-JEPA~\cite{bardes2023vjepa}. They are both trained on masked videos -- VideoMAE is tasked with predicting the pixels, while V-JEPA predicts features encoded by a teacher network. Both models use a ViT backbone and tokenize videos into patches of size (2, 16, 16), applying self-attention between all of the tokens. We evaluate different variants of VideoMAEv1, VideoMAEv2 and V-JEPA, ranging from ViT-B with 30M parameters to ViT-g with 1B parameters which are, to the best of our knowledge, the largest that exist.

\begin{figure*}
    \centering
    \small{
    \begin{subfigure}[b]{\textwidth}
        \centering
        \begin{tabular}{@{}p{0.125\textwidth}@{}p{0.125\textwidth}@{}p{0.125\textwidth}@{}p{0.125\textwidth}@{}p{0.125\textwidth}@{}p{0.125\textwidth}@{}p{0.125\textwidth}@{}p{0.125\textwidth}@{}@{}}
            \centering 20M & \centering 100M & \centering 300M & \centering 600M & \centering 2B & \centering 4B & \centering Frame & \centering Masked frame 
        \end{tabular}
        {\includegraphics[width=\textwidth]{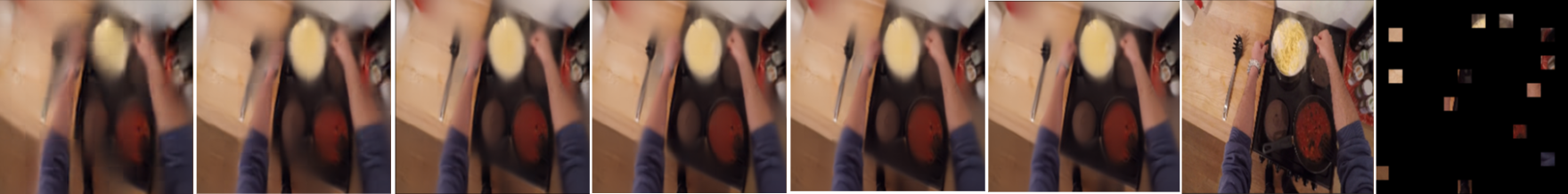}}\\[0.2em]
        {\includegraphics[width=\textwidth]{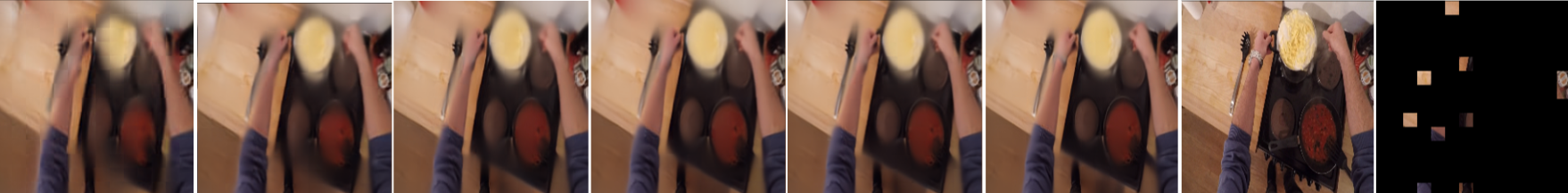}}
        \caption{Reconstructions of the 1st (\textbf{top}) and 9th (\textbf{bottom}) frames (0.64s apart) from a 16-frame Epic Kitchens \cite{Damen2018EPICKITCHENS} video using \Model models of different sizes (the 22B model is omitted due to differing input/patch dimensions). The last two columns show the original and 95\% randomly masked frame.}
        \label{subfig:epic_20M_4B}
        \vspace{0.8em}
    \end{subfigure}
    
    \begin{subfigure}[b]{\textwidth}
        \centering
        \begin{tabular}{@{}p{0.111\textwidth}@{}p{0.111\textwidth}@{}p{0.111\textwidth}@{}p{0.111\textwidth}@{}p{0.111\textwidth}@{}p{0.111\textwidth}@{}p{0.111\textwidth}@{}p{0.111\textwidth}@{}p{0.111\textwidth}@{}}
            \centering 20M & \centering 100M & \centering 300M & \centering 600M & \centering 2B & \centering 4B & \centering 22B & \centering Ground truth & \centering Frame
        \end{tabular}
        \scalebox{1}[1]{\includegraphics[width=\textwidth]{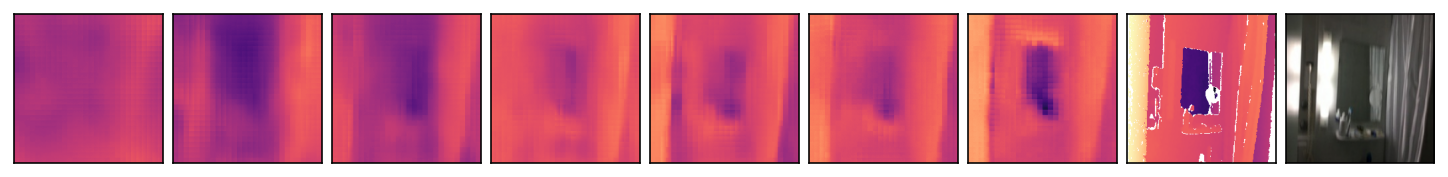}}\\[-0.4em]
        \caption{Depth map predictions on the ScanNet dataset. White regions in the ground truth indicate missing depth information.}
        \label{subfig:scannet_20M_20B}
        \vspace{0.8em}
    \end{subfigure}

    \begin{subfigure}[b]{\textwidth}
        \centering
        \begin{tabular}{@{}p{0,142\textwidth}@{}p{0,142\textwidth}@{}p{0,142\textwidth}@{}p{0,142\textwidth}@{}p{0,142\textwidth}@{}p{0,142\textwidth}@{}p{0,142\textwidth}@{}}
            \centering 20M & \centering 100M & \centering 300M & \centering 600M & \centering 2B & \centering 4B & \centering 22B
        \end{tabular}
        \scalebox{-1}[1]{\includegraphics[width=\textwidth]{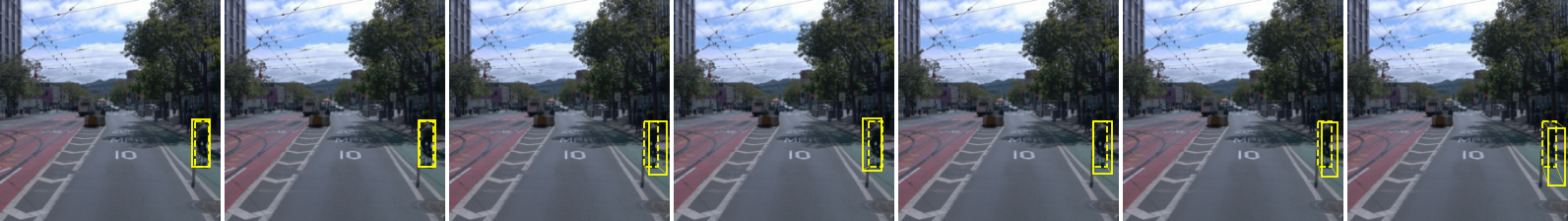}}\\[0.2em]
        \caption{Object tracking on Waymo dataset. Ground truth boxes are depicted with dashed lines, while predicted boxes use solid lines.}
        \label{subfig:waymo_20M_20B}
    \end{subfigure}
    \caption{Qualitative results of \Model models (20M-22B). We first illustrate MAE training: masked and reconstructed frames on Epic Kitchens (a). We then assess downstream task performance: depth prediction on ScanNet (b) and object tracking on Waymo (c).}
    \label{fig:scaling_epic_scannet_waymo_20M_20B}
    }
\end{figure*}

\begin{figure}[t]
    \centering
    \includegraphics[width=\columnwidth]{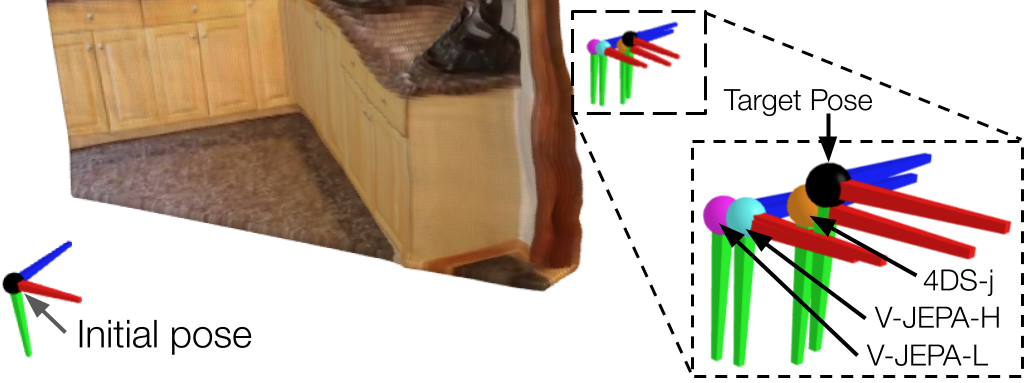}
    \caption{Pose estimation qualitative results on RealEstate10k. Given the features extracted from a small video fragment of 16 frames, we obtain video features with our pretrained video model backbones, and train a pair-wise pose predictor readout to predict the target pose. Improvement with model scaling is observed.}
    \vspace{-3mm}
    \label{fig:posere10k}
\end{figure}

\begin{figure*}
    \centering
    \includegraphics[width=0.95\linewidth]{img/readout_depth_performance_v4.png}
    \vspace{-1mm}
    \caption{The performance of the frozen \Model-j model when attaching  attention-based readouts at different layers of the model (as a percentage of the total number of self-attention blocks, 64 for \Model-j as given in~\cref{tab:vit_configs}). The layer at 95\% depth (60 for \Model-j) offers a good compromise across tasks -- we report results using this readout layer throughout the paper except for action classification frozen evaluation where we use layer 75\%.\vspace{-2mm}}
    \label{fig:readout_depth_performance}
\end{figure*}

\section{Results}
\label{sec:results}

Our first experiments aimed to understand how the strong models listed in~\cref{sec:baselines} fare on the 4D representation tasks described in~\cref{sec:evals} under the frozen evaluation protocol, with the same trainable attention-based readouts, same limited amount of training. Note that some of the results are lower than in papers without these constraints, which train longer (for example for SSv2), do test-time augmentation or use specialized readouts; this is to be expected and is by design. The results, provided in the next subsection, highlighted VideoMAE and V-JEPA as robust approaches; we pursued MAE because of its better training efficiency (it uses a fixed rate of token dropping, which is more friendly on TPUs / with batching and does not require an EMA network), for the \Model models.

\subsection{Assessing video models 4D representations}

 RE10k camera pose estimation uses Mean End Point Error (explained in supp. material, lower is better and minimum value is 0); Perception test point tracking uses Average Jaccard (higher is better, range 0-1); ScanNet depth prediction uses Absolute Relative Error (lower is better, minimum value is 0) and Waymo Open object tracking uses mean IoU (higher is better, range 0-1). Results for SSv2 and Kinetics700-2020 action classification are reported by Top-1. We report accuracy and Average Jaccard in tables as percentages for better readability.

\Cref{tab:baselines} shows the quantitative results of the selected models: image models in first three rows, then existing video models, and finally our \Model models at the bottom.

\vspace{2mm}
\noindent \textbf{Image backbones.} An interesting initial finding was that image backbones perform poorly regardless of model size on all 4D tasks with the exception of depth estimation on ScanNet -- likely because monocular cues dominate in short clips just over 1s long, where parallax is naturally limited and also where there are no moving objects. Dino does less poorly than SigLIP across all tasks, hinting that  CLIP-style representations may not be a good fit for 4D tasks. All models are terrible at camera pose estimation, but they are good at Kinetics, which is known to depend strongly on the quality of high-level (semantic) appearance features.

\vspace{2mm}
\noindent \textbf{Video backbones.} Text-supervised VideoPrism models show strong performance on depth estimation, but are not as good on the tracking tasks and are quite poor on camera pose estimation. V-JEPA models exhibit overall strong performance. Compared to the smaller v1, VideoMAE-v2 uses additional language-pretraining which seems to lead to worse performance on the 4D tasks. Language pretrained models do however very well on the semantic tasks, SSv2 and Kinetics700-2020. VJEPA does great on SSv2.

\vspace{2mm}
\noindent \textbf{\Model backbones.} Our scaled up models, \Model-G, e and j -- explained in~\cref{sec:simplemae} -- significantly improve over all the baselines, achieving top results on all 4D tasks. On semantic tasks they are not as competitive (although results continuously improve with scale), illustrating why prior self-supervised video learning work, which has focused on those tasks, has not pursued scaling. The smaller 4DS models are not as good as corresponding VideoMAE models. Partly this may be because 4DS model features are loaded from the layer at 95\% depth, so from slightly shallower features and without an extra decoder, as well as the other simplifications we introduced.

\vspace{2mm}
\noindent \textbf{Distilling from large models.} We  tried distilling a 4DS-e (4B parameters) into a B model (90M parameters). Results are shown in~\cref{tab:baselines} and suggest that one can also obtain SOTA small models by distilling from the larger ones. In detail -- we input the same video to both the randomly initialized 4DS-B model and the pretrained 4DS-e frozen model, without dropping any patches. We trained the 12 layer B model to approximate the activations of two different layers of the teacher model, for 100k steps using batch size 1024 and the L1 loss.  We added two parallel linear layers at the end of the B model: one predicted teacher layer 36 activations, the other one layer 51 activations (note 4DS-e has many more channels than B, so adapter layers were unavoidable); afterwards we discarded these extra linear layers -- they were only used during distillation. The improvement over 4DS-B model trained by itself is quite noticeable and raises interesting questions for future work.

\begin{figure*}[t]
    \centering
    \includegraphics[width=\linewidth]{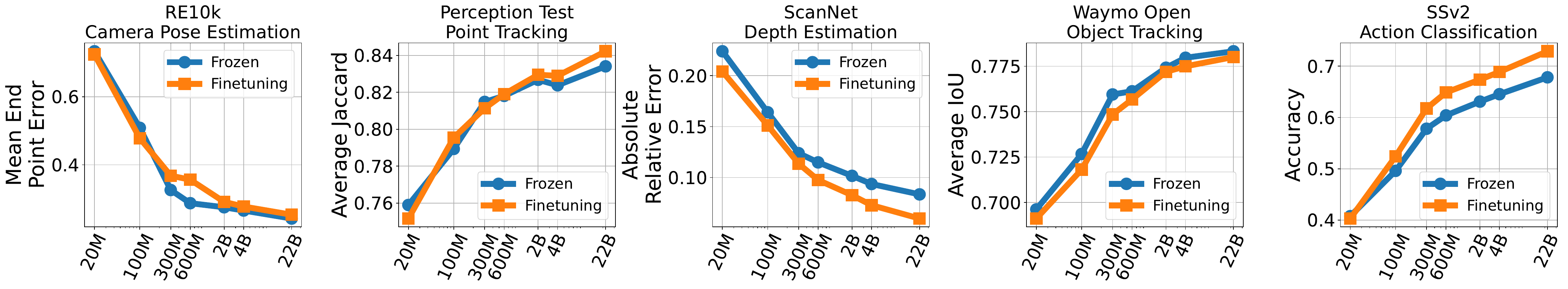}
    \vspace{-5mm}
    \caption{Model performance with increasing model sizes ranging from 20M parameters to 22B parameters, on all five tasks, with  frozen evaluation and short finetuning.  Model size is plotted on log-scale, at points \Model-S (20M), \Model-B (100M), \Model-L (300M), \Model-H (600M), \Model-G (2B), \Model-e (4B) and \Model-j (22B).\vspace{-3mm}}
    \label{fig:scalingall}
\end{figure*}

\vspace{2mm}
\noindent \textbf{Finetuning evaluation}. We repeated the evaluation using end-to-end finetuning -- exactly the same as before but training both readout and backbone (instead of freezing the backbone). Note that this is significantly more expensive and also that we did not perform any test-time augmentation nor additional train-time augmentations relative to the frozen evaluation. Results for a short finetuning schedule of 20k steps are shown in~\cref{tab:baselines_ft} and are consistent with the frozen evaluation results. There are big improvements over frozen evaluation in ScanNet and SSv2, on  others tasks results are comparable. Results with longer schedules are reported in~\cref{tab:baselines_longer_ft} in the supplementary material. Frozen and finetuning performance across \Model model sizes is summarized visually in~\cref{fig:scalingall}. In all cases performance continues to improve with model size.

\subsection{Other studies}

\noindent \textbf{Which layer is best for each task?} As mentioned in the technical section, for \Model we did not force a clear separation between encoder and decoder (except for a patchwise linear decoder). We investigated  where in the model are the best representations for each task by trying a subset of 6 readout layers for each \Model model at different model depths -- 25\%, 50\%, 75\%, 85\%, 95\% and 100\% (e.g.\ 75\% corresponds to layer 42 out of 56 for \Model-e).~\cref{fig:readout_depth_performance} shows that for most tasks the layer at 95\% depth seems like a good option. For the semantic tasks (SSv2, Kinetics), we used the layer at 75\% of model depth for frozen evaluation; for finetuning we also used 95\% except for Kinetics where we used 75\%. These are the readout layers we used throughout the paper to report results.

\vspace{2mm}
\noindent \textbf{Data scaling impact.} We also evaluated the value of the amount of training. Results for training over 250M, 500M and 1B examples for model \Model-e is shown in~\cref{tab:num_examples}. There are nearly always gains from extra training. %

\vspace{2mm}
\noindent \textbf{Decoding hyperparameters}. We include an ablation confirming our defaults in the supp. material, sec.~\ref{sec:decoding_study}.

\begin{table}

\centering
\begin{tabular}{lrrrr}
\toprule
\small \# examples & 250M & 500M & 1B \\
\midrule
\small RE10k & 0.35 & 0.29 & \textbf{0.27} \\
\small Perception Test & 81.8 & \textbf{82.9} & 82.4 \\
\small ScanNet & 1.39 & 1.11 & \textbf{0.94} \\
\small Waymo Open & 74.7 & 76.8 & \textbf{78.0} \\
\small SSv2 & 59.0 & 64.3 & \textbf{68.2} \\
\bottomrule
\end{tabular}
\vspace{-1mm}
\caption{Impact of number of examples seen in pretraining on downstream performance for an 4DS-e model (4B parameters).\label{tab:num_examples}\vspace{-4mm}}
\end{table}

\subsection{Qualitative results}

To further help understand the impact of model size, we visualize the outputs of our models during MAE training in \cref{subfig:epic_20M_4B} and on depth estimation in \cref{subfig:scannet_20M_20B}, object tracking in \cref{subfig:waymo_20M_20B} and camera pose estimation in \cref{fig:posere10k}. All obtained using a frozen backbone.

Larger models consistently demonstrate superior performance. During MAE training, they exhibit improved reconstruction quality: in \cref{subfig:epic_20M_4B}, the reconstruction of the spatula becomes progressively sharper as model size increases. This trend holds true for downstream tasks as well. In depth estimation, smaller models produce washed out, inaccurate results. As model size increases, details emerge (see \cref{subfig:scannet_20M_20B}). Camera pose estimation results are illustrated in \cref{fig:posere10k} (larger models also tend to be more accurate).

\section{Conclusions}
\label{sec:conclusions}

We have re-evaluated several state-of-the-art image and video models, trained with various types of supervision including language, on a set of mostly semantic-free tasks requiring strong 4D representations. We found that many of the improvements found in the respective papers do not translate well to these more geometrical / temporal scene understanding tasks.

We then demonstrate that scaling MAE beyond what has been done in the literature ($\sim$1B parameters), all the way to  22B parameter models brings consistent improvement -- we have a new collection of model checkpoints we call \Model which may be useful for the community. The largest models may not be practical for all applications, but it is likely they can be distilled to smaller form factors while preserving much of their performance. Generally, we hope that the scaling insights in this paper  may open new avenues for further progress in vision similar to those that have been explored successfully in language models research.

\vspace{2mm}
\noindent \textbf{Limitations}. In this paper we used the most vanilla version possible of MAE (which we called SimpleMAE) -- many improvements have been reported in the literature at small scale. As future work it would be interesting to  identify which of these variations are most practical and fruitful when scaling, or if other learning approaches are superior. Another limitation of the paper is that we did not produce "scaling laws" -- but arguably provided evidence that this may soon be warranted.

\vspace{2mm}
\noindent \textbf{Acknowledgments} We would like to thank several people. Relja Arandjelovic for reviewing a draft of the paper and proposing several improvements. Rahul Sukthankar and Kevin Murphy for valuable feedback and pointers. David Bridson and Mirgahney Mohamed for helping with data infrastructure. Jon Scholz, Yulia Rubanova, Alex Bewley, Lili Momeni, Nikhil Parthasarathy, Maks Ovsjanikov, Shiry Ginosar, Alexey Dosovitskiy, Alexander Lerchner, Tengda Han, Catalin Ionescu, Yusuf Aytar and Stannis Zhou for helpful discussions. Guy Scully and Junlin Zhang for great program management. Satinder Baveja and Raia Hadsell for supporting the project.

{
    \small
    \bibliographystyle{ieeenat_fullname}
    \bibliography{main}
}
\clearpage
\setcounter{page}{1}
\maketitlesupplementary

\appendix

This supplementary material provides additional details on pretraining, evaluation protocol, tasks, readouts and baselines. It also includes additional qualitative results about SSv2, ScanNet and Waymo Open evaluations as well as extra MAE pretraining visualizations and a small study on correlation between the 5 tasks, when taking into account results from all baselines and \Model models. Finally, we have an ablation study on decoding hyperparameters and a table with long end-to-end finetuning results for all models.

\section{Additional details}
\label{sec:evals}

\subsection{Pretraining}

Pretraining used slightly different settings for the 6 \Model models up to 4B parameters (small) and for the 22B parameters model (large). Details for both cases are listed in~\cref{table:pretraining}. Training longer and using slightly larger patch size for the 22B model were deliberate choices. The other parameters were picked either by necessity (to avoid running out of memory with the 22B model -- number of decoding layers and output patch size) or based on short runs as it would be too expensive to tune them exhaustively -- the 22B model took 20 days to complete training\footnote{We trained a 22B model using much larger batch size, 8192, and it trained significantly faster (1 week) due to better usage of accelerator parallelism, but results were inferior (perhaps with tuning it will work as well).}.
 As data augmentation we used only random cropping (spatially and temporally) and left-right flipping. For optimization, we used AdamW, omitting from weight decay the learned positional embeddings and layer norm parameters.

\begin{table*}[t]
\centering
\begin{tabular}{c c c}
\hline
\textbf{Hyperparameter} & \textbf{20M-4B models} & \textbf{22B model} \\
\hline
\# Training steps & 488,282 & 1,000,000 \\
Input resolution & 224$\times$224 & 256$\times$256 \\
Learning rate & 1e-4 & 2e-4 \\
Warmup steps & 10,000 & 100,000 \\
\# N decoding layers & 4 & 2 \\
\# Output patch size & 2$\times$16$\times$16 & 4$\times$32$\times$32 \\
Minimum resize factor & \multicolumn{2}{c}{1.15} \\ 
Input patch size & \multicolumn{2}{c}{2$\times$16$\times$16} \\ 
Batch size & \multicolumn{2}{c}{2048} \\
MAE masking ratio & \multicolumn{2}{c}{0.95} \\ 
AdamW weight decay & \multicolumn{2}{c}{0.05} \\
AdamW b1 & \multicolumn{2}{c}{0.90} \\ 
AdamW b2 & \multicolumn{2}{c}{0.95} \\ 
\hline
\end{tabular}
\caption{Pretraining hyperparameters for 20M-4B (left) and for 22B (right) \Model models. The nature of some of them may not be immediately obvious -- minimum resize factor dictates how much videos minimum side gets resized as function of input resolution (e.g. 224x224 gets randomly cropped from videos resized to 1.15*224 smallest side).}
\label{table:pretraining}
\end{table*}

\subsection{Evaluation}

For frozen evaluation, we train models for any of the tasks using the same number of training examples, 1.28M, by doing as many epochs over the training dataset as necessary. Batch size is fixed in all cases to 32, which means models train for 40k steps. We use the AdamW optimizer, with weight decay 1e-4 and sweeping learning rate (1e-4, 3e-4, 1e-3) then taking the best result. We start the training with linear warmup for 1k steps, from a 0.0 learning rate, then use cosine decay all the way down to 1e-7.

Finetuned evaluation follows a similar protocol to the frozen one, however, has a few different settings. While we sweep the learning rate for frozen evaluation, we sweep learning rate (1e-4, 3e-4) and weight decay (1e-4, 5e-2). Besides, the linear warmup is for 3k steps during finetuning instead of 1k. We report the results for short finetuning (20k steps) in the main paper and, medium (40k steps) and long finetuning (80k steps) in the supplementary material.

For all tasks, we use the same cross-attention readout architecture but with different hyperparameters (e.g. number of channels) and queries for each task (explained in detail in their respective sections below). The readout takes in the spatio-temporal features from the frozen encoder and cross-attends to these using  learned or positional queries to predict the task output (following the setup in \cite{bardes2023vjepa}). In more detail, we first pass features through layer normalization and then add learned temporal embeddings. Input queries attend to these features to output task features that are then passed through a residual MLP with a hidden size of 4 times the size of task features. A final linear layer then maps the task features to the desired output size for the task.

The rest of this subsection will go in detail into each of the 5 tasks in our evaluation, the task specific parameterization of readouts done in each case and current top results in literature where available. A summary of the readout configurations and number of parameters in each case is provided in~\cref{table:readout_modules}.

\begin{table*}[t]
\centering
\begin{tabular}{c|l c}
\hline
\textbf{Eval} & \textbf{Architecture} & \textbf{Number of Params} \\ \hline
SSv2 Classification & 
\begin{tabular}[c]{@{}l@{}}
\texttt{LearnedQueries(num\_channels=768)} \\
\\
\texttt{CrossAttention(qkv\_size=768, num\_heads=12)} \\
\\ 
\texttt{Linear(output\_size=174)}
\end{tabular} & 7,041,966 \\ \hline
K700-2020 Classification & 
\begin{tabular}[c]{@{}l@{}}
\texttt{LearnedQueries(num\_channels=1024)} \\
\\
\texttt{CrossAttention(qkv\_size=1024, num\_heads=16)} \\
\\ 
\texttt{Linear(output\_size=700)}
\end{tabular} & 12,281,532 \\ \hline
RE10k Pose Prediction & 
\begin{tabular}[c]{@{}l@{}}
\texttt{LearnedQueries(num\_channels=256)} \\
\\
\texttt{CrossAttention(qkv\_size=256, num\_heads=8)} \\
\\
\texttt{Linear(output\_size=12)}
\end{tabular}  & 1,650,444 \\ \hline
ScanNet Depth Prediction & 
\begin{tabular}[c]{@{}l@{}}
\texttt{LearnedQueries(num\_channels=1024)} \\
\\
\texttt{CrossAttention(qkv\_size=1024, num\_heads=16)} \\
\\
\texttt{Linear(output\_size=128)}
\end{tabular} & 18,116,736 \\ \hline
Waymo Open Object Tracking & 
\begin{tabular}[c]{@{}l@{}}
\texttt{FourierQueries(num\_bases=16)} \\
\\
\texttt{MLP(hidden\_size=512, output\_size=512)} \\
\\
\texttt{CrossAttention(qkv\_size=1024, num\_heads=4)} \\
\\
\texttt{Linear(output\_size=4)}
\end{tabular} & 12,482,624 \\ \hline
Perception Test Point Tracking  & 
\begin{tabular}[c]{@{}l@{}}
\texttt{FourierQueries(num\_bases=16)} \\
\\
\texttt{MLP(hidden\_size=512, output\_size=512)} \\
\\
\texttt{CrossAttention(qkv\_size=1024, num\_heads=8)} \\
\\
\texttt{Linear(output\_size=4)}
\end{tabular} & 12,396,552 \\ \hline
\end{tabular}
\caption{Configurations and number of parameters of cross-attention-based readout modules used in this paper, for different tasks. Note that the number of parameters are given for the case of a ViT-L backbone, that has 1024-channel outputs.}
\label{table:readout_modules}
\end{table*}

\subsubsection{Action classification 1: Something-Something v2}
The SSv2 action classification dataset contains 220,000 videos with duration ranging from 2 to 6 seconds at 12fps. Videos contain 174 human actions with everyday objects. It is designed for fine-grained understanding of human hand gestures like putting something into something and turning something upside down. 

\vspace{2mm}
\noindent \textbf{Task definition.} Given a video clip of 16 frames of resolution 224x224 with stride 2, the model is tasked to predict an action class. Top-1 accuracy is used to measure the performance.

\begin{table}[t]
\centering
\begin{tabular}{l|l}
\hline
 Brightness & delta in [-0.125, 0.125] \\ \hline
 Saturation & factor in [0.6, 1.4] \\  \hline
 Contrast & factor in [0.6, 1.4] \\  \hline
 Hue & delta in [-0.2, 0.2]  \\ \hline
\end{tabular}
\caption{Hyper-parameters for color augmentation used when training readout heads on the SSv2 task. Deltas are added to the corresponding channel, while factors multiply the corresponding channel.}
\label{tab:colour_augment}
\end{table}

\vspace{2mm}
\noindent \textbf{Readout details.} The cross-attention readout module uses 768 channels with 12 heads and a single learned query to predict logits for 174 classes. In training, we resize the shorter size of the video to 239 and take random temporal crop of shape 224x224 from it. We use colour augmentation with 0.8 probability of randomly adjusting the brightness, saturation, contrast and hue (see \cref{tab:colour_augment}), and a 0.1 probability of converting to grayscale. In test time we take one 224x224 central crop from the video without any colour augmentation.

\vspace{2mm}
\noindent \textbf{Current SOTA.} V-JEPA-ViT-H~\cite{bardes2023vjepa} achieves 68.5\% through attention probing by training on 2M video clips. When finetuning results can be as high as 77.3~\cite{wang2023masked} but using significant volume of train and test-time augmentations that are complex to replicate. In this paper we use in all cases a single clip at test time and moderate/reasonable amounts of training augmentation.

\subsubsection{Action classification 2: Kinetics-700-2020}

\vspace{2mm}
\noindent \textbf{Task definition.} Given a video clip of 16 frames of resolution 224x224 with stride 2, the model is tasked to predict one action class out of 700. At test time, since videos are variable length and up to 10s long (250 total frames at 25 fps), we first loop the video as many times as necessary to get 250 frames, then take 7 linearly spaced 16-frame clips and pass them through the model. Finally, we average the softmax scores across all of them. Top-1 accuracy is used to measure the performance.

\vspace{2mm}
\noindent \textbf{Readout details.} The same readout is used as for SSv2, but slightly larger -- 1024 channels and 16 heads instead of 768 and 12.

\vspace{2mm}
\noindent \textbf{Current SOTA.} Vision-language models are very strong in Kinetics. Among them, VideoPrism~\cite{zhao2024videoprism} is one of the strongest models for frozen attention-based evaluation on kinetics-400 (an earlier, smaller version of the Kinetics-700-2020 version we use). It is also the strongest one in our evaluation.

\subsubsection{Relative pose estimation: RealEstate10k}

We adopt the RealEstate10k dataset~\cite{realestate10k,zhou2018stereo} which consists of a collection of videos of properties obtained from YouTube, annotated with camera poses using a traditional structure-from-motion (SfM) pipeline. To resolve the scale ambiguity inherent to traditional SfM methods, we employ the technique described in \cite{watson2024controlling} for re-scaling the camera poses to metric units. We adopt the original splits defined in the dataset, which contain around 10 million training frames coming from 6.5k training videos, and around 1 million test frames coming from 696 test videos. 

\vspace{2mm}
\noindent \textbf{Task definition.} We test the ability of learnt video representations to encode 6DoF relative camera poses. To this end, given a sequence of N video frames and their corresponding representations, we train a readout head to predict the relative pose between the first and last frames of the sequence, given their learnt representations.

\vspace{2mm}
\noindent \textbf{Evaluation metrics.} We evaluate the quality of relative poses using a metric based on end-point-error (EPE) that looks at the rotation and translation components jointly. This metric measures the mean distance of points transformed with the ground-truth ($P_i$) and estimated ($\hat{P}_i$) poses: $e_{\text{EPE}}(\hat{P}_i, P_i) = \frac{1}{N}\sum_{j=1}^N \| P_i(X_j) - \hat{P}_i(X_j) \|$. Here, $\{X_j\}_{j=1,\dots,N}$ corresponds to a set of points in 3D space where we want to evaluate the metric. In practice, we employ 8 auxiliary points forming a virtual cube in front of the camera of the first frame for computing $e_{\text{EPE}}$.

\vspace{2mm}
\noindent \textbf{Readout details.} We concatenate the features for the first and last frame of the input video and feed this to a cross-attention readout head with channels parameters, 8 heads and a single learned query to output a $12-$dimensional vector corresponding to the matrix form of an SE(3) pose transformation, including a $3\times3$ estimated rotation matrix and a $3\times1$ translation vector. Because the estimated rotation matrix predicted by the model is not guaranteed to respect the properties of a true rotation matrix in SO(3), we apply the Procrustes method~\cite{bregier2021deep} to project it to the closest true rotation matrix before computing the metrics. The readout is trained by minimizing the L2-loss between the predicted and ground-truth pose matrices. 

\vspace{2mm}
\noindent \textbf{Current SOTA.} This is a new combination of dataset / task / metric we created as part of the research in this paper. As a strong baseline, zero shot DUSt3R~\cite{wang2024dust3r} gets EPE of 1.06, which is worse than our model (0.24). This is because Dust3r sometimes fails to register the two pointmaps  producing  large  errors.

\subsubsection{Point tracking: Perception Test}

Perception Test \citep{patraucean2023perception} is a multimodal benchmark of real world videos that contains various tasks to evaluate the perception and reasoning skills of multimodal models. Here we focus on the point tracking task and evaluate on the validation set that consists of 73 videos with an average length of 722 frames. Each video contains multiple point tracks annotated by human annotators following the protocol of TAP-Vid \citep{doersch2022tapvid}. On average, each video contains 60 point tracks where a point is visible on average in 480 frames.

For training the point tracking readout head, we use the Kubric MOVi-E dataset \citep{greff2022kubric}, which consists of synthetic videos of scenes containing 10-20 static objects and 1-3 dynamic objects rendered on a high-resolution photo as background. The camera moves on a straight line with random constant velocity and is always pointed towards the origin of the scene. The training set consists of 97500 24-frame videos with point annotations computed as described in \citep{doersch2022tapvid}.

\vspace{2mm}
\noindent \textbf{Task definition.} The aim of the point tracking task is to produce the position $x$, $y$ of a query point over multiple frames given the video and an initial query point position. To adapt the pretrained models for point tracking, we append a cross-attention readout head that takes in the video features calculated by the pretrained model and a set of query points, and outputs a prediction for each frame and point track. Following the setup in \cite{doersch2023tapir}, this prediction consists of $x$, $y$ position of the point, whether it is visible, and uncertainty (i.e., how confident the model is in its position prediction). We train the readout head (and the backbone if finetuning) using a weighted sum of 3 losses: Huber loss over position predictions, binary cross entropy over visibility and uncertainty with weights 100, 0.1, and 0.1 respectively (see \citep{doersch2023tapir} for more details).

For training, we randomly sample 16 frames and 64 point tracks from each video. Then we select a random crop with an area range of (0.3, 1.0) and aspect ratio range of (0.5, 2.0) and resize the video to $224\times 224$. Query points are always chosen from the first frame, and any point tracks where the query point is not visible in the first frame are ignored. We train for 40K steps with a batch size of 32 using learning rate of 3e-4 and a cosine decay schedule with 1000 warm-up steps.

For evaluation, we again pick query points from the first frame in a video and ignore any tracks where the query point is not visible in the first frame. We evaluate on clips of 16 frames sampled with a stride of 4.

\vspace{2mm}
\noindent \textbf{Evaluation metrics.} To evaluate each model, we report the Average Jaccard (AJ) metric \citep{doersch2022tapvid}, which evaluates both occlusion and position accuracy, and is an average of Jaccard values for thresholds of 1, 2, 4, 8, 16 pixels. Given a threshold, the Jaccard metric calculates the fraction of `true positives', i.e., points within the threshold of any visible ground truth points, divided by `true positives' plus `false positives' (points that are predicted visible, but the ground truth is either occluded or farther than the threshold) plus `false negatives' (ground truth visible points that are predicted as occluded or the prediction is farther than the threshold). 

\vspace{2mm}
\noindent \textbf{Readout details.} We again use the same cross-attention readout architecture used in other tasks but. Since we have query points for each track, we do not learn queries for cross-attention but use the query point positions as queries. We found it is useful to embed these query point positions using Fourier features (with 16 bases) and pass them through an MLP with 512 hidden and output units before they are fed to cross-attention as queries. We use cross-attention with 1024 parameters and 8 heads. Instead of predicting position for all frames from a single query, we found it is better to replicate the query for a track 8 times (adding learnable temporal embeddings) and predict 2 frames at a time from each query. The output of the readout head for each frame and point track is a 4D vector predicting point position, visibility, and uncertainty. We found having a sigmoid activation on position outputs to improve results.

\vspace{2mm}
\noindent \textbf{Current SOTA.} BootsTAPIR \cite{doersch2024bootstap} is a specialized model for point tracking that is trained end-to-end for this task -- it reports 59.6\% AJ on full videos with query first sampling. When evaluated on 16 frames, this model gets 92.5\% AJ. Note that BootsTAPIR is already trained on point tracking so there is no need to train a new readout head.

\subsubsection{Object tracking: Waymo Open}

The Waymo Open Dataset \citep{9156973} consists of high resolution (1280x1920) RGB videos and LiDAR data captured at 10fps using the hardware on Waymo cars in urban and suburban environments. These videos are annotated with 2D and 3D bounding boxes. We use only the RGB data as input to our models and the 2D bounding boxes to compute the loss and metrics. The dataset contains 798 training samples and 202 validation samples which are 20s long.

\noindent \textbf{Task definition.} We train a readout head to predict the position $x_{min}$, $x_{max}$, $y_{min}$, $y_{max}$ of query boxes over the 16 frame clip given the video and the position of boxes in the initial frame. 
For both the training and evaluation, following prior work \cite{steenkiste2024moving, elsayed2022savi, kipf2022conditional} we downsample the video to 256x384 and the fps to 5, we then take a central crop of 256x256 and a random temporal crop of 16 frames. We filter out any boxes which occupy $<$ 0.5\% of the first sampled frame and limit the number of boxes to a maximum of 25.

\noindent \textbf{Evaluation metrics.} 
We report the IoU (intersection over union) averaged over boxes and over frames excluding the initial frame where box positions are given. We compute this over 4096 samples from the validation set, note that, although the validation set contains only 202 clips, we take random temporal crops leading to a higher number of samples.

\vspace{2mm}
\noindent \textbf{Readout details.} The readout head mostly follows the setup used for the point tracking task, with query box positions embedded using Fourier features and used as queries for cross-attention. We use cross-attention with 1024 channels and 4 heads and do not any apply activation function on the output. For object tracking, we found predicting all frames from a single query for each box track works well, so we do not replicate the queries with learned temporal embeddings as we have done for point tracking.

\vspace{2mm}
\noindent \textbf{Current SOTA.}
Object tracking results reported on this dataset vary in terms of the resolution, and crucially the number of frames evaluated over and the maximum number of boxes evaluated per clip. Furthermore, evaluations are not computed on the exact same clips and boxes due to temporal cropping in evaluation and filtering of boxes. As a comparison point, MooG~\cite{steenkiste2024moving} reports IoU of 0.73 with a frozen backbone using the same cropping and filtering at a resolution of 128x128 (while we use 256x256). This uses a readout head with a different recurrent architecture; and it is trained on the Kubric MOVi-E dataset\citep{greff2022kubric} and evaluated 0-shot on Waymo Open dataset. However this is on the easier problem of tracking up to 10 boxes over 8 frames while we evaluate tracking up to 25 boxes over 16 frames. 

When trained end-to-end on Waymo Open dataset and evaluated over 10 frames and up to 10 boxes, MooG~\cite{steenkiste2024moving} reports IoU of 0.67. SAVi++~\cite{elsayed2022savi} in the same setting reports 0.68. We report 0.83 for the best model when tracking over 16 frames up to 25 boxes.

\subsubsection{Depth estimation: ScanNet}

ScanNet~\cite{dai2017scannet} is a video dataset captured in various indoor environments, containing rich annotations for 3D camera poses, surface reconstructions, and instance-level semantic segmentations. Data was obtained through an RGB-D capture system that produces depth. RGB frames have 1296x968 resolution while depth frames have 640x480 resolution. There are 1201 videos in the train split, 312 videos in the validation split, and 100 videos in the test split. We use the train and validation splits of ScanNet for this paper.

\noindent \textbf{Task definition.} We feed the models 16 RGB frames while adding readout heads on top to output depth for each input frame. We scale the images to the (0, 1) range and mask out target depth values outside of (0.001, 10) meters. We perform random cropping and left-right flipping during training and take a center crop during testing. 

\noindent \textbf{Evaluation metrics.} We follow prior work on monocular depth estimation and report the mean of the absolute relative error (AbsRel) \citep{steenkiste2024moving,yang2024depth,ranftl2021vision} which is computed as $|d^{*}-d|/(d+\epsilon)$ where $d^{*}$ is the predicted depth values, $d$ is the ground truth depth. 

\noindent \textbf{Readout details.} We use a cross-attention readout head with 1024 parameters and 16 heads with one learned query for each spatio-temporal patch in the input video. We use a patch size of $2 \times 8 \times 8$ and predict 128 ($=2*8*8$) depth values for each patch, one for each pixel. We use an L2 loss.

\noindent \textbf{Current SOTA.} Some of the most recent methods include ZoeDepth \cite{bhat2023zoedepth} and Depth Anything \cite{yang2024depth}. Neither evaluates the method on ScanNet videos and they all evaluate zero-shot, which makes it hard to compare our work with these methods apples-to-apples. Among these, DUSt3R~\cite{wang2024dust3r} zero-shot gets 0.088 whereas a slightly older method~\cite{wang2022less} gets 0.093. An earlier work that trains and evaluates on ScanNet videos is Atlas~\cite{murez2020atlas} which takes known camera intrinsics and extrinsics and achieves 0.061 AbsRel. Our best numbers with frozen features are 0.084 and with full finetuning 0.057 (we do not use camera parameters).

\subsection{Baselines}

\subsubsection{SigLIP-2B}

We evaluate the 2B parameter image model used as part of PaLI-3~\cite{chen2023pali}, which builds on the SigLIP~\cite{zhai2023sigmoid} family of Vision Transformers~\cite{dosovitskiy2020image}.

\vspace{2mm}
\noindent \textbf{Architecture.} This model consists of 48 self-attention pre-norm transformer layers, with 1536 feature dimension, and 16 attention heads. The transformer feed-forward block has a hidden dimension of 8192.

\noindent \textbf{Model size.}  2 billion parameters.

\noindent \textbf{Training objective.} Given image-text pairs, a vision embedding and a text embedding are independently computed. A binary classifier then estimates for all pairs of image-text embedding dot products, whether they belong together or not. See the SigLIP paper~\cite{zhai2023sigmoid} for more details on the training dataset and loss. Only the vision tower is used in our evaluations, the text encoder is discarded.

\subsubsection{VideoPrism} 

VideoPrism is a video encoder that performs well on a range of video understanding tasks.

\vspace{2mm}
\noindent \textbf{Architecture.} VideoPrism is based on ViViT architecture which has a factorized design in space and time. However, VideoPrism excludes the global average pooling after the spatial encoder for an improvement on the downstream tasks with fine-grained details.

\noindent \textbf{Model size.} They experiment two sizes: VideoPrism-g and VideoPrism-B which are based on ViT-giant with 1B parameters and ViT-Base network.

\noindent \textbf{Training objective.} The training of VideoPrism consists of two stages. Initially, the network is trained on video-text pairs contrastively. The objective at this stage is to minimize a symmetric cross-entropy loss over the similarity scores of all video-text pairs. The second stage is masked video modeling which is done to improve the "motion" understanding of the model as the first stage focuses on "appearance". This stage only uses video frames. The model at this stage (student) is trained to predict embeddings from masked frames similar to the predictions of the first stage model (teacher) from unmasked frames. They also introduce strategies like token shuffling and global distillation losses at this stage.

\noindent \textbf{Pretraining details.} Pretraining corpus includes 612M clips from various sources including public datasets like WTS-70M, YT-Temporal-180M, and InternVid as well as some non-public datasets. They train for $5 \times 10^{5}$ iterations (with additional $4.5 \times 10^{4}$ iterations for warmup).

\subsubsection{V-JEPA} 

\vspace{2mm}
\noindent \textbf{Architecture.} V-JEPA uses a vision transformer\cite{dosovitskiy2020image} as encoder.  It takes 16 video frames with resolution (224, 224) as input, and embeds each video patch of shape (2, 16, 16) into a latent space using a linear projection. These embeddings are then augmented with positional encodings and fed into self-attention blocks, where all the patches self-attend to each other.

\noindent \textbf{Model size.} We evaluate two variants of the models, ViT-L(~300M parameters) and ViT-H(~600M parameters). We used the official model checkpoint from \footnote{https://github.com/facebookresearch/jepa}.

\noindent \textbf{Training objective.} In joint-embedding predictive architecture (JEPA), the model is trained to predict the representation of an input y from the representation of another input x, where y is the original input video and x is the same video but with large number of patches dropped. In the pre-training stage, there are two encoders $E_x(x)$ and $E_y(y)$  which process input x and y respectively. The L1 distance between the output of $E_x(x)$ is optimized to match the the one from  $E_y(y)$ after a projection module. $E_x(x)$ is updated by standard back-propagation and $E_y(y)$ is updated as an exponential moving average of the weights of $E_x(x)$. In downstream evaluation, only $E_y(y)$ is used as the encoder.

\noindent \textbf{Pretraining details.} Pretraining is done on VideoMix2M which combines videos from HowTo100M, Kinetics-400/600/700 (K710)~\cite{kay2017kineticshumanactionvideo} and  Something-Something-v2~\cite{goyal2017something}. The model is trained for 90,000 iterations with batch size 3072.

\subsubsection{DINOv2}

\noindent \textbf{Architecture.} The DINOv2 image encoder is a standard vision transformer~\cite{oquab2023dinov2, dosovitskiy2020image}, with patch size 14. It takes in 16 video frames with resolution (224, 224), and independently encodes each video frame, producing a grid of 16x16x16 feature vectors. The output feature vector dimensionality depends on the ViT variant; we use two ViT sizes, the ViT-L and ViT-G, with 1024 and 1536 hidden dimensions, respectively. The two encoder sizes have 307M and 1.1B parameters, respectively.

\noindent \textbf{Training objective.} We re-use the pre-trained checkpoint provided by~\cite{oquab2023dinov2}. The main losses used during pretraining include a self-distillation loss, masking loss, and various centering and regularization techniques. Contrastive and distillation losses are applied on both the whole-image-aggregate and patch levels. Pretraining image data curation also plays a central role for the model, and involves rounds of data duplication and diversification to ensure wide coverage of many concepts. More details about the pretraining and pretraining dataset can be found in~\citep{oquab2023dinov2}.

\noindent \textbf{Pretraining details.} The model checkpoint used was pretrained for 625K steps, with a 100K warmup, and cosine decay LR schedule starting at 3e-4. The pretraining batch size, as reported by the paper, is 3K images per batch.

\section{Additional qualitative results}

We include additional qualitative results obtained by models with frozen features and learned attention readouts on top.

\begin{figure*}
    \centering
    \includegraphics[width=\linewidth]{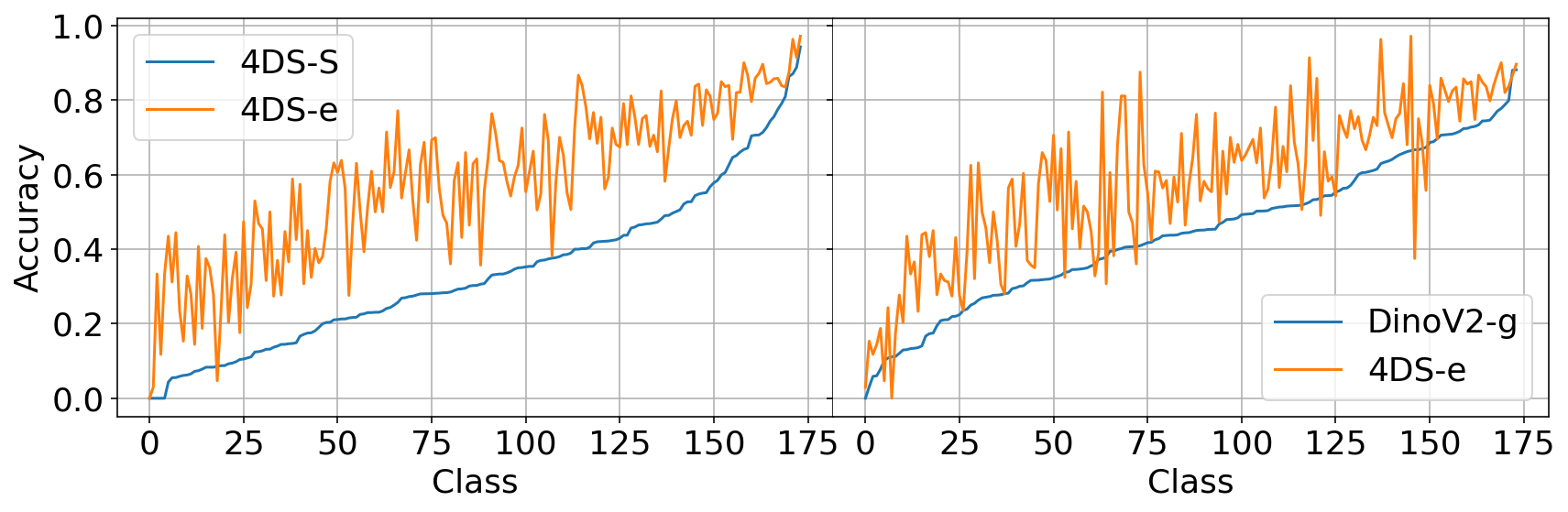}
    \includegraphics[width=\linewidth]{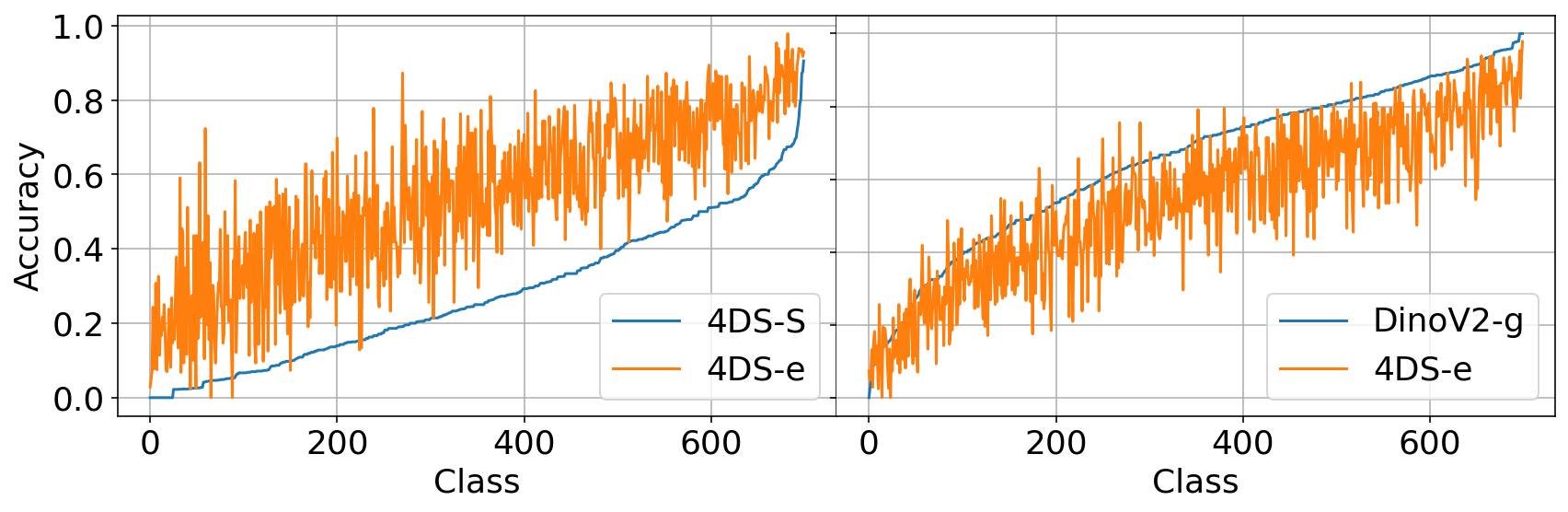}
    \caption{Top-1 accuracy of the 20M parameter \Model-S (blue) vs the 4B parameter \Model-e model (orange) on the SSv2 (above) and Kinetics700-2020 (below) datasets and the DinoV2-g image model vs the \Model-e model. Classes are ordered by the accuracy of the model shown in blue.}
    \label{fig:ssv2_accuracy_comparisons}
\end{figure*}

\noindent \textbf{SSv2 action classification.} Per-class accuracy differences are visualized for the SSv2 action classification task comparing a large \Model model to a small one and to DinoV2-g in~\cref{fig:ssv2_accuracy_comparisons}. The only class where accuracy gets worse when comparing \Model-S and \Model-e is ``Taking something from somewhere" which both models perform badly on, getting between 4\% and 9\%. 

\begin{figure*}
    \centering
    \begin{tabular}{@{}p{0.111\textwidth}@{}p{0.111\textwidth}@{}p{0.111\textwidth}@{}p{0.111\textwidth}@{}p{0.111\textwidth}@{}p{0.111\textwidth}@{}p{0.111\textwidth}@{}p{0.111\textwidth}@{}p{0.111\textwidth}@{}}
        \centering 20M & \centering 100M & \centering 300M & \centering 600M & \centering 2B & \centering 4B & \centering 22B & \centering Ground truth & \centering Frame
    \end{tabular}
    \scalebox{1}[1]{\includegraphics[width=\textwidth]{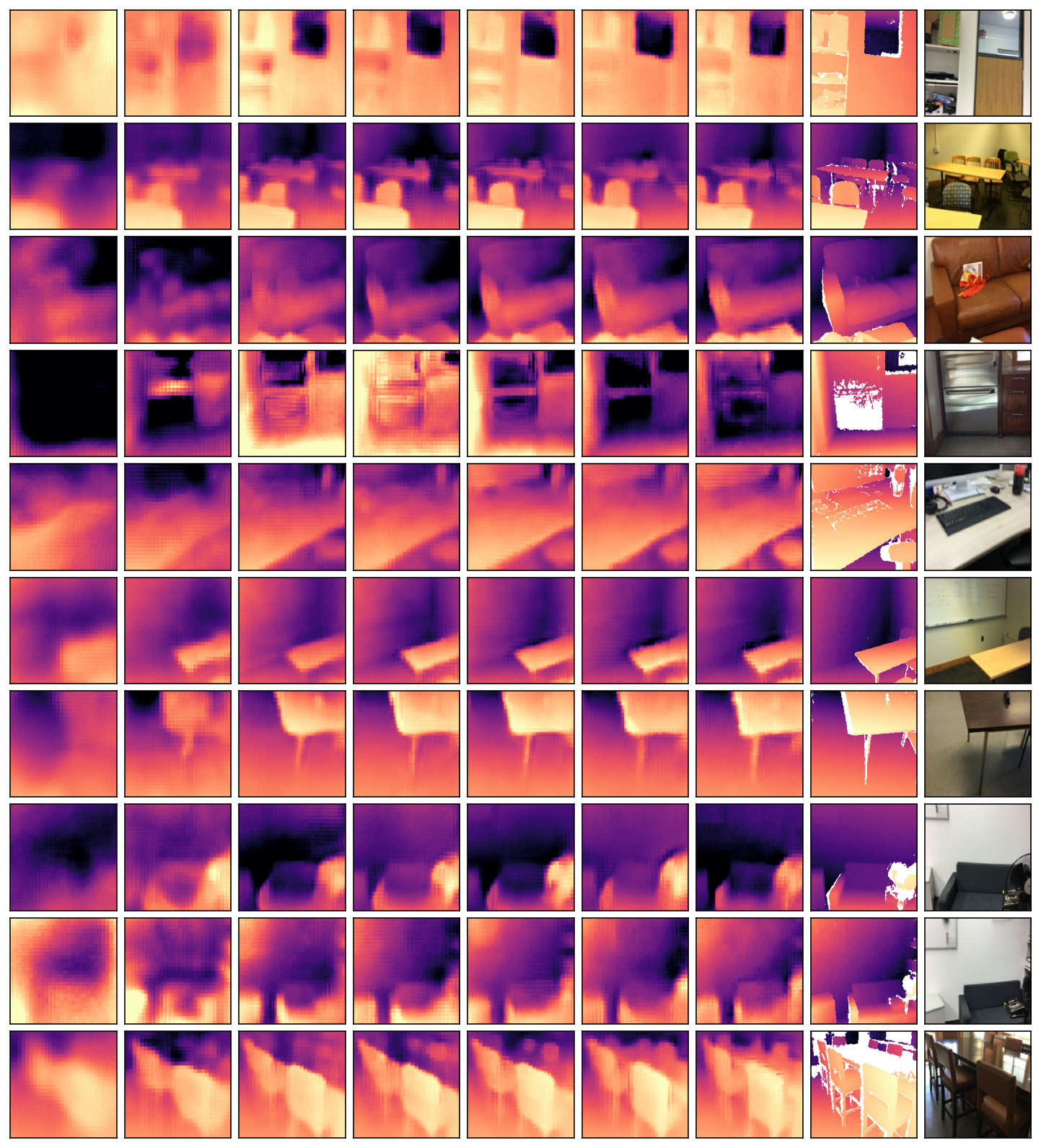}}
    \caption{Comparison of depth map predictions from MAE models with varying parameter sizes (20M to 22B) on the ScanNet dataset.  Each row displays the predicted depth map for the first frame of a video sequence, alongside the corresponding ground truth and RGB image. White regions in the ground truth indicate missing depth information.}
    \label{fig:scaling_scannet_20M_20B_more}
\end{figure*}

\vspace{2mm}
\noindent \textbf{ScanNet depth prediction.} Additional qualitative results for ScanNet depth prediction are provided in~\cref{fig:scaling_scannet_20M_20B_more}.

\begin{figure*}
    \centering
    \begin{tabular}{@{}p{0,142\textwidth}@{}p{0,142\textwidth}@{}p{0,142\textwidth}@{}p{0,142\textwidth}@{}p{0,142\textwidth}@{}p{0,142\textwidth}@{}p{0,142\textwidth}@{}}
        \centering 20M & \centering 100M & \centering 300M & \centering 600M & \centering 2B & \centering 4B & \centering 22B
    \end{tabular}
    \scalebox{-1}[1]{\includegraphics[width=\textwidth]{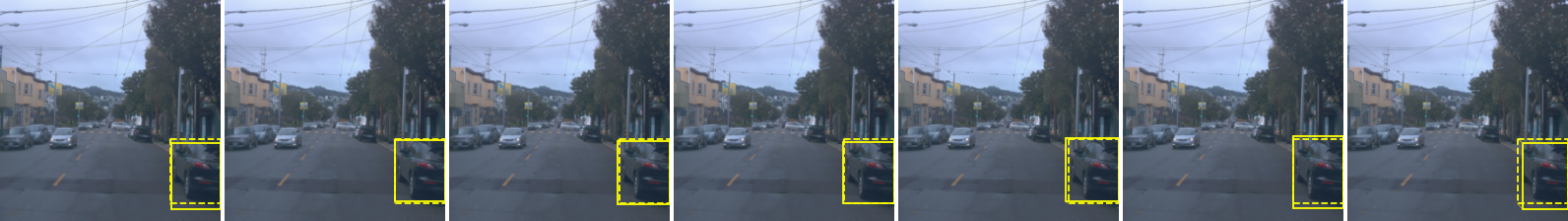}}\\[0.2em]
    \scalebox{-1}[1]{\includegraphics[width=\textwidth]{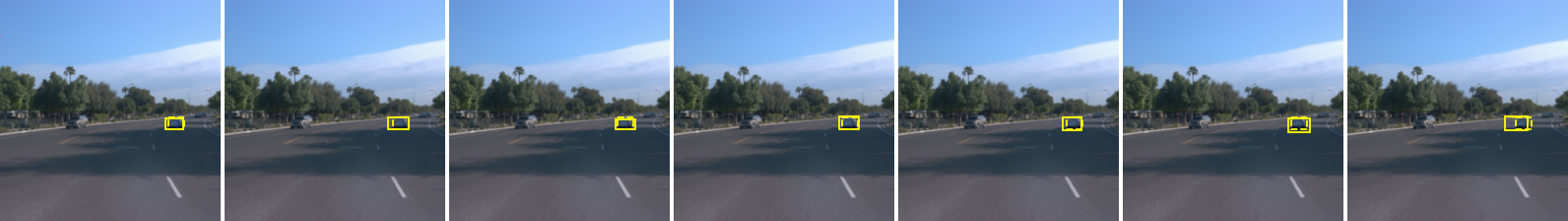}}\\[0.2em]
    \scalebox{-1}[1]{\includegraphics[width=\textwidth]{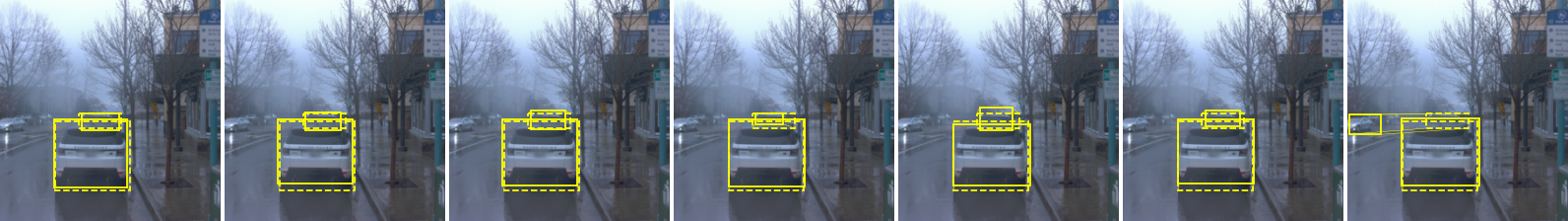}}\\[0.2em]
    \scalebox{-1}[1]{\includegraphics[width=\textwidth]{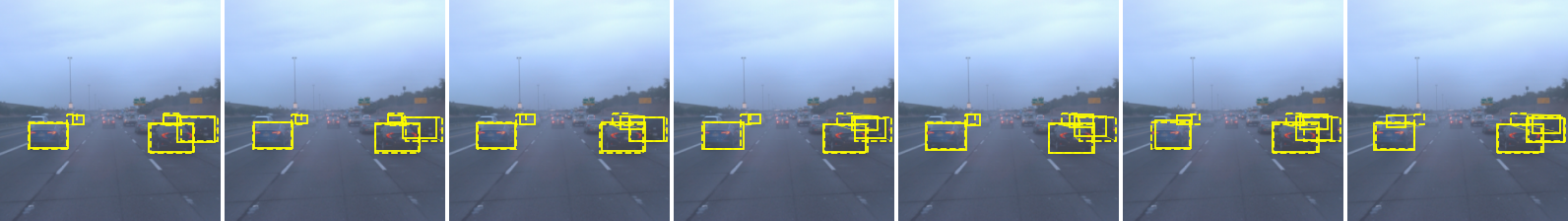}}\\[0.2em]
    \scalebox{-1}[1]{\includegraphics[width=\textwidth]{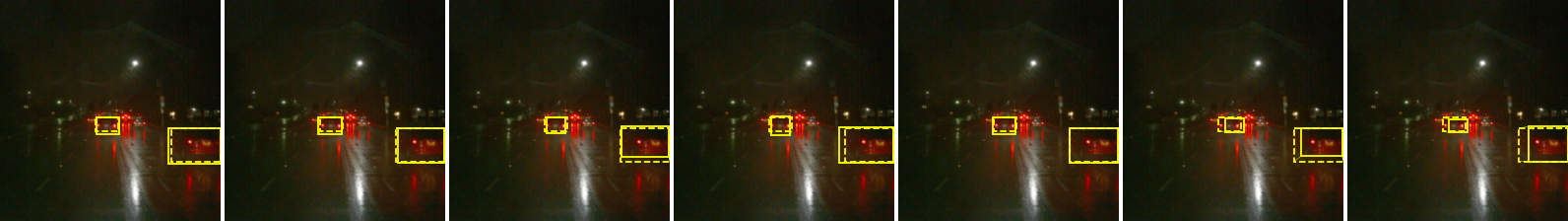}}\\[0.2em]
    \scalebox{-1}[1]{\includegraphics[width=\textwidth]{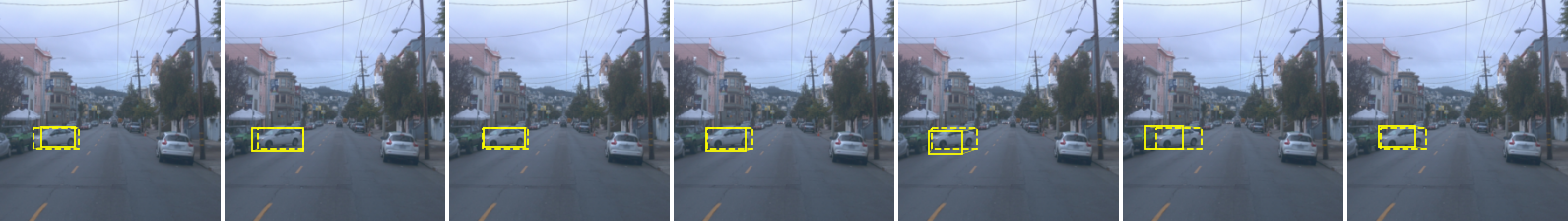}}\\[0.2em]
    \scalebox{-1}[1]{\includegraphics[width=\textwidth]{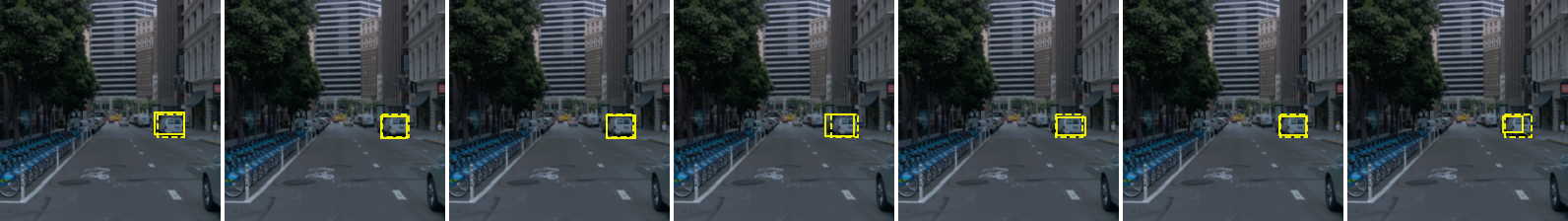}}\\[0.2em]
    \scalebox{-1}[1]{\includegraphics[width=\textwidth]{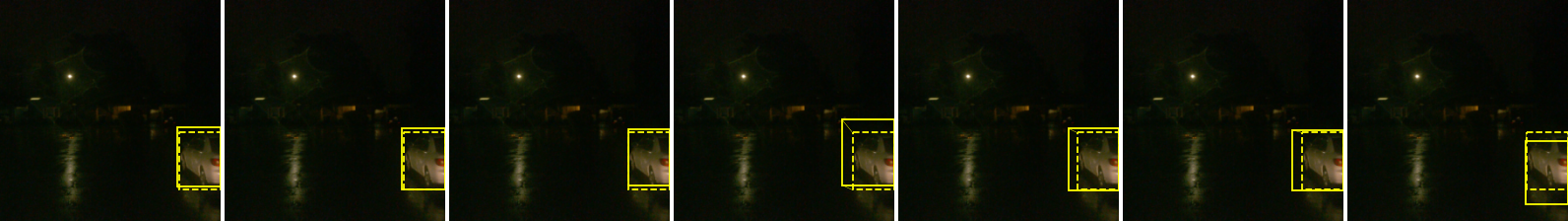}}
    \caption{Visualization of predicted and ground truth bounding boxes for object tracking on the Waymo Open Dataset.  Results are shown for the final frame of different video sequences using various MAE model sizes. Ground truth boxes are depicted with dashed lines, while predicted boxes use solid lines.}
    \label{fig:scaling_waymo_20M_20B_more}
\end{figure*}

\vspace{2mm}
\noindent \textbf{Waymo Open object tracking.} Additional qualitative results for Waymo Open object tracking are provided in~\cref{fig:scaling_waymo_20M_20B_more}.

\vspace{2mm}
\noindent \textbf{Perception Test point tracking.} Additional qualitative results for Perception Test point tracking are provided in~\begin{figure*}
\centering
\begin{tabular}{@{}p{0.5\textwidth}@{}p{0.5\textwidth}@{}}
\centering \Model-S (20M) & \centering \Model-j (22B)
\end{tabular}
\scalebox{-1}[1]{\includegraphics[width=\columnwidth]{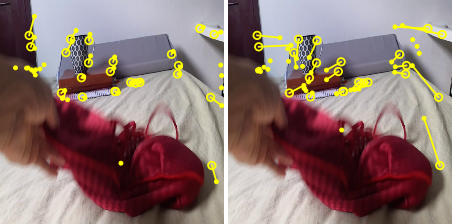}}
\scalebox{-1}[1]{\includegraphics[width=\columnwidth]{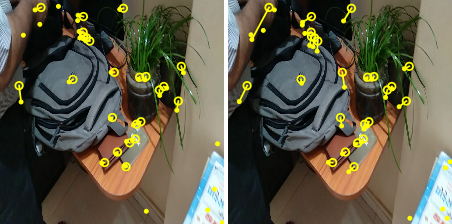}}\\[0.2em]
\caption{Point tracking error visualization for 20M (left) and 22B parameter (right) \Model models on the Perception Test dataset. We show the final frame for different videos. Large circles represent ground truth point locations, smaller ones represent predicted locations --  lines connecting the two indicate error vectors. The larger model seems more competent at this correspondence task, particularly for points in less textured areas of the scene.}
\label{fig:scaling_perception_test_20M_20B}
\end{figure*}
. Larger models significantly reduce errors, visualized as displacement errors, particularly around hands and moving objects.

\begin{figure*}
    \centering
    \begin{tabular}{@{}p{0.125\textwidth}@{}p{0.125\textwidth}@{}p{0.125\textwidth}@{}p{0.125\textwidth}@{}p{0.125\textwidth}@{}p{0.125\textwidth}@{}p{0.125\textwidth}@{}p{0.125\textwidth}@{}@{}}
        \centering 20M & \centering 100M & \centering 300M & \centering 600M & \centering 2B & \centering 4B & \centering Frame & \centering Masked Frame 
    \end{tabular}
    {\includegraphics[width=\textwidth]{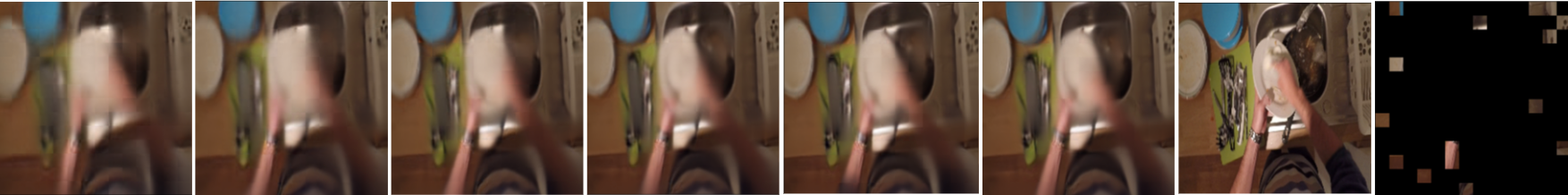}}\\[0.2em]
    {\includegraphics[width=\textwidth]{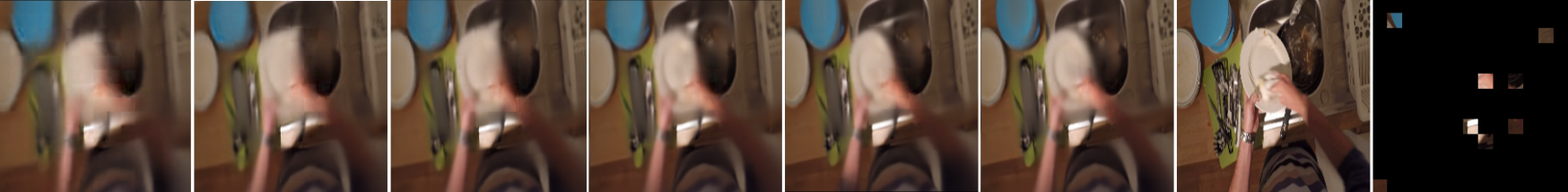}}\\[0.2em]
    {\includegraphics[width=\textwidth]{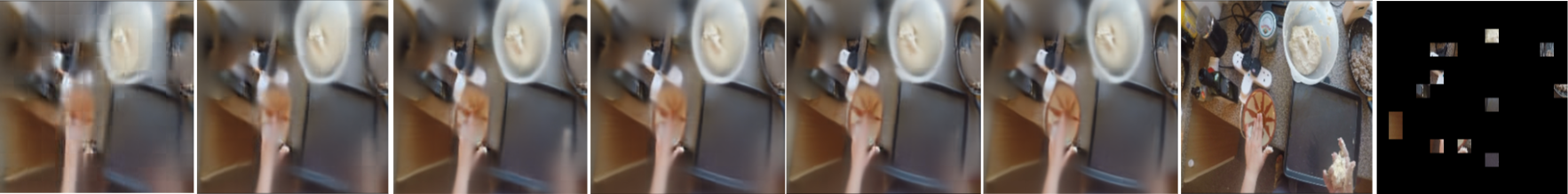}}\\[0.2em]
    {\includegraphics[width=\textwidth]{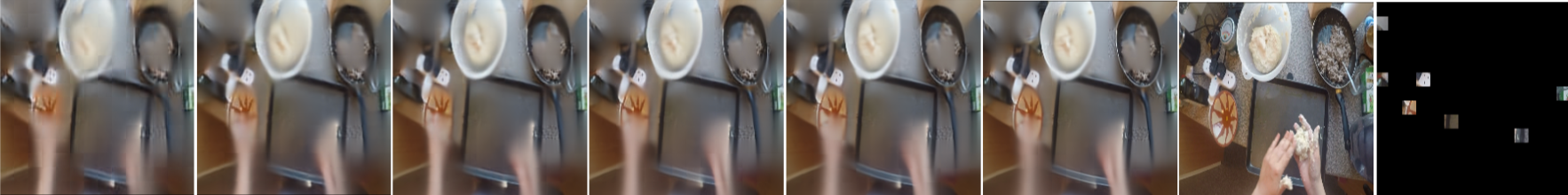}}\\[0.2em]
    \caption{Reconstructed frames on the Epic Kitchens \cite{Damen2018EPICKITCHENS} dataset from MAE models of increasing size. Each column labeled with a model size shows a reconstruction of the 1st and 9th frames (0.64s apart) of two different 16 frame videos given as input to the model. The original frame and this same frame with 95\% random masking prior to reconstruction are shown in the final two columns. The 22B parameter model is omitted since it uses 256x256 crops and therefore does not have compatible masking with the other models.}
    \label{fig:mae_reconstruct}
\end{figure*}

\vspace{2mm}
\noindent \textbf{MAE reconstruction results.} Additional qualitative results on the pretraining task of masked auto-encoding are provided in~\cref{fig:mae_reconstruct}. This visualization uses videos from the Epic Kitchens dataset.

\begin{figure*}
    \centering
    \includegraphics[width=\linewidth]{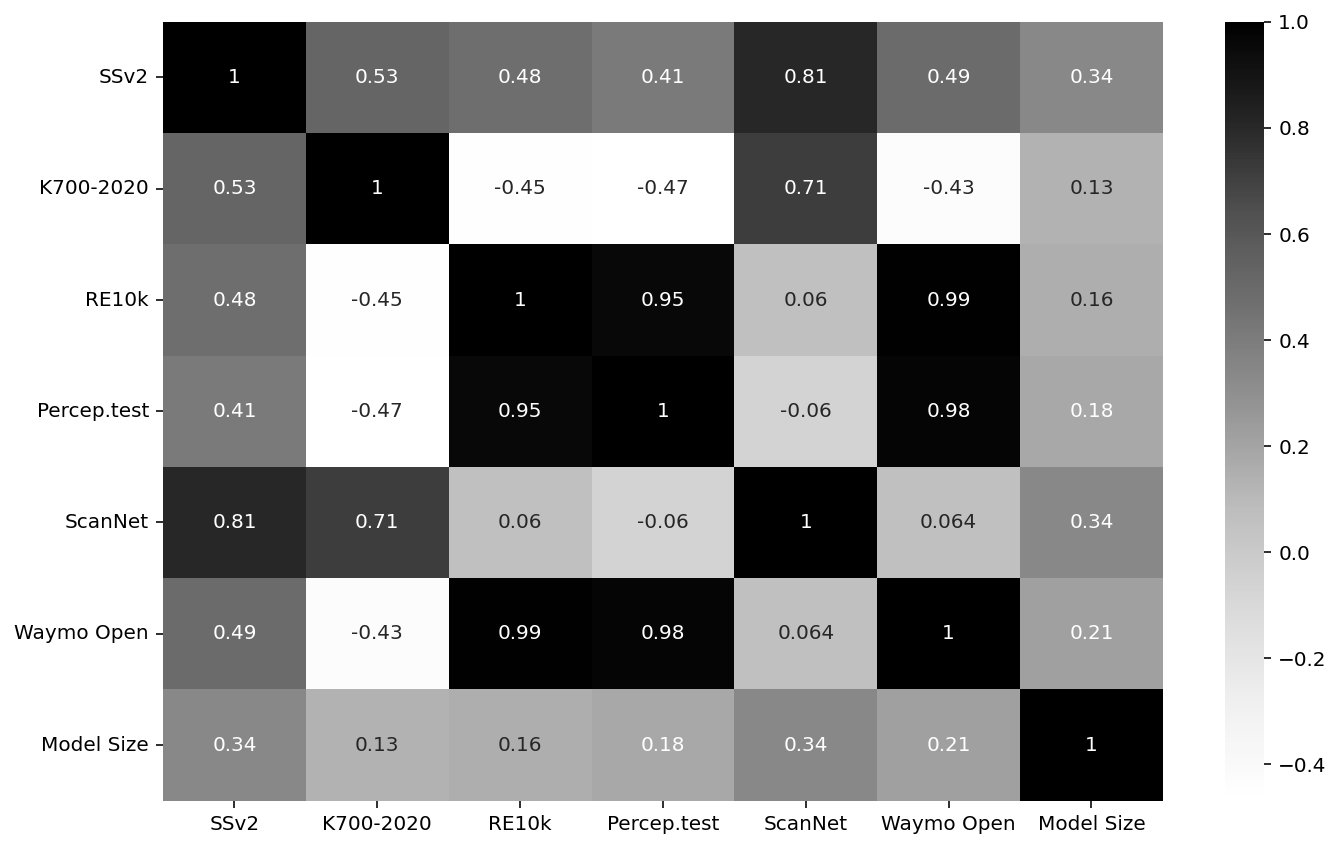}
    \caption{Pearson's correlation coefficients for all frozen attention-based evaluation tasks across all models. This visualization is based on the results provided by \cref{tab:baselines}. The RE10k and ScanNet values have been inverted, so that larger values correspond to better performance. A highly semantic task, K700-2020, displays negative correlation with most other non-semantic tasks. Model size has a positive correlation with all tasks suggesting that larger model size leads to better performance.}
    \label{fig:correlation}
\end{figure*}

\vspace{2mm}
\noindent \textbf{Task correlation.}
The correlation coefficient for all evaluation tasks are provided in~\cref{fig:correlation}. There is a positive correlation with model size suggesting that larger model size leads to better performance. Kinetics classification is the least correlated task. 

\begin{figure*}
    \centering
    \includegraphics[width=1.3\columnwidth]{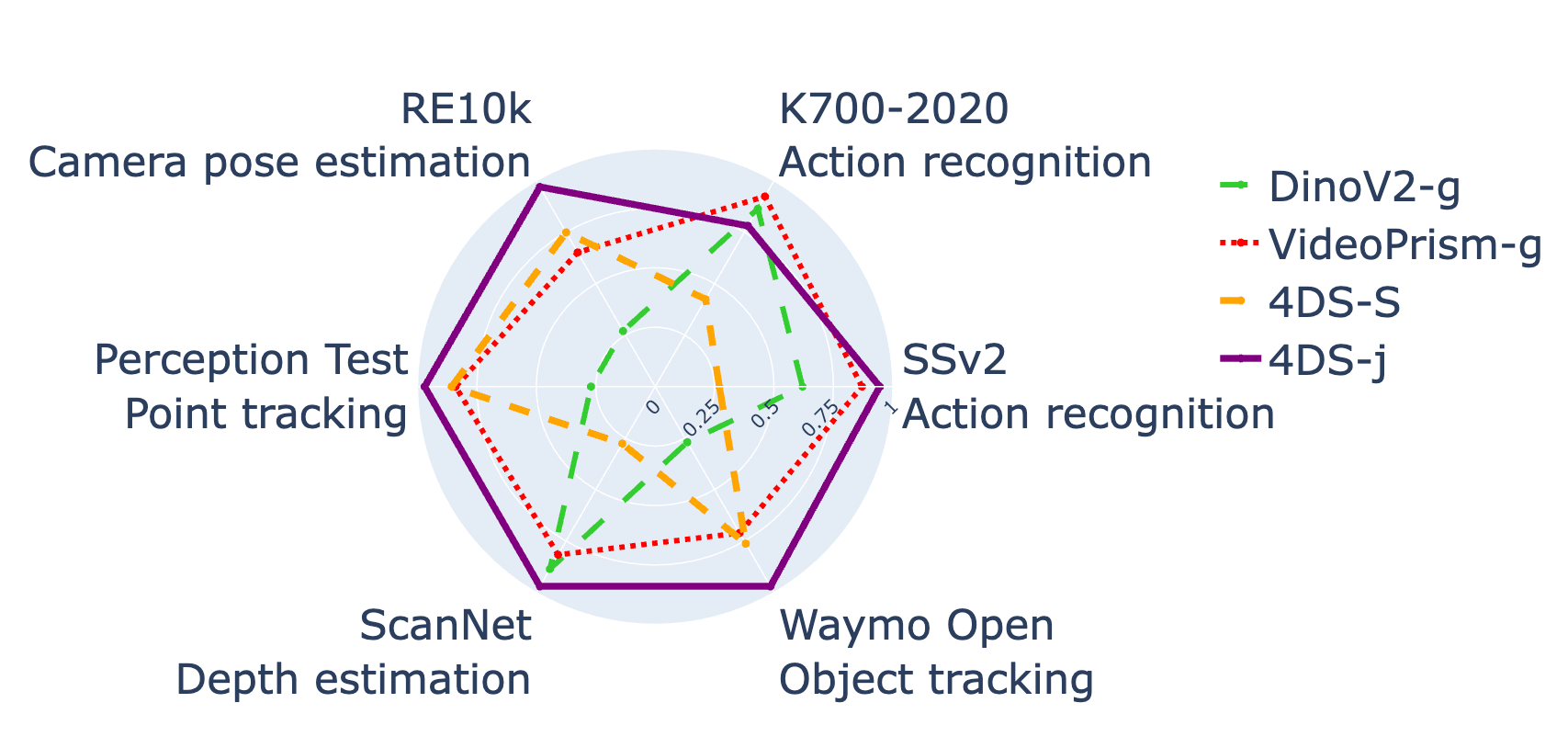}
    \caption{Radar plot comparing the frozen evaluation performance on all tasks of the image model DinoV2-g, video model VideoPrism-g, and 2 \Model video models of different sizes, 20M (\Model-S) and 22B parameters (\Model-j). Note that we rescale the ScanNet Absolute Relative Error and RE10k Mean End Point Error using a sigmoid function so that 100 is 0 error and 0 is infinite error. We further scale all values for visualization purposes between 0.3 and 1 where 1 is the best result among these models and 0.3 corresponds to the worst result. The image encoder has obvious representational deficiencies on temporal tasks (point and object tracking, camera pose). VideoPrism-g, a video model with text-focused training is more well-rounded but on the more temporal tasks is still slightly inferior to even the smallest 4DS model.}
    \label{fig:radar_plot_scale_and_im}
\end{figure*}

\vspace{2mm}
\noindent \textbf{Additional relative performance visualization.} A radar plot showing relative performance visually between some of the strongest models in our comparison is shown in~\cref{fig:radar_plot_scale_and_im}.

\section{Decoding hyperparameters study}
\label{sec:decoding_study}
There are two main hyperparameters in our approach to decoding in SimpleMAE: the decoder patch size and the number of latent layers (where latent tokens self-attend with regular tokens). Results for \Model-e are shown in~\ref{tab:decoding_hypers} and validate the default values we used for all models but \Model-j -- which due to memory constraints uses 2 latent layers and patch size (4,32,32). Presumably with larger accelerators or a more efficient implementation it would be possible to get extra performance out of this model using the same hyperparameters as for all other models.

\begin{table}[htbp]
\centering
\caption{Study on decoding hyperparameters using the RE10k, ScanNet and SSv2 tasks, pre-training the \Model-e model on 1B examples (with quick evaluation on 4096 test examples for each task). 4 latent layers and (2,16,16) patches seems preferable in all cases.}
\label{tab:decoding_hypers}
\setlength{\tabcolsep}{9pt}
\begin{tabular}{llcc}
\toprule
\multirow{2}{*}{\textbf{Task}} & \multirow{2}{*}{\textbf{Decoding patch size}} & \multicolumn{2}{c}{\textbf{\# latent layers}} \\ 
 &  & \multicolumn{1}{c}{\textbf{2}} & \multicolumn{1}{c}{\textbf{4}} \\ \midrule
\multirow{2}{*}{\textbf{RE10k}} & (2,16,16) & \multicolumn{1}{c}{0.32} & \textbf{0.29} \\  
 & (4,32,32) & \multicolumn{1}{c}{0.32} & 0.33 \\ \hline
\multirow{2}{*}{\textbf{ScanNet}} & (2,16,16) & \multicolumn{1}{c}{1.0} & \textbf{0.95} \\ 
 & (4,32,32) & \multicolumn{1}{c}{1.15} & 1.12 \\ \hline
\multirow{2}{*}{\textbf{SSv2}} & (2,16,16) & \multicolumn{1}{c}{63.3} & \textbf{65.8} \\ 
 & (4,32,32) & \multicolumn{1}{c}{60.6} & 61.9 \\ 
 \bottomrule
\end{tabular}
\end{table}

\section{Extra Finetuning Results}
\label{sec:finetuning}

In the main paper we presented frozen evaluation and short-schedule finetuning evaluation (20k steps). Here we include also results for medium and long-schedule finetuning evaluation -- 40k and 80k steps in~\cref{tab:baselines_longer_ft}.
Note again that we did not perform any test-time augmentation nor additional train-time augmentations compared to the frozen evaluation. As a practical detail, we found it useful to use a lower learning rate for the backbone compared to the readout. We scale down the backbone learning rate relative to the readout by a factor 0.003.

\begin{table*}[t]
\centering
\caption{Comparison of existing (first two blocks of rows) and new models (\Model, last set of rows) with \textbf{medium} and \textbf{long finetuning} -- 40k and 80k steps respectively (nothing is frozen). Results are displayed as medium / long finetuning. All results are obtained using the same trainable readouts -- see text for details. ScanNet values are multiplied by 10 for better readability. Bold highlights the best result in each task.}
\setlength{\tabcolsep}{3pt}
\begin{tabular}{l@{\hskip -0.08in}c@{\hskip 0.05in}c@{\hskip 0.05in}|cc|cccc}
\toprule
\textbf{Model} & Size (M) & \multirow{2}{*}{\parbox{1.6cm}{\centering Image (I)/ Video (V)}} & \textbf{SSv2} & \textbf{K700-2020} & \textbf{RE10k} & \multirow{2}{*}{\parbox{1.6cm}{\centering \textbf{Percep. test ($\uparrow$)}}} & \textbf{ScanNet} & \multirow{2}{*}{\parbox{1.6cm}{\centering \textbf{Waymo Open ($\uparrow$)}}} \\ 
& &  &($\uparrow$) &($\uparrow$)&($\downarrow$)& &($\downarrow$)& \\
\midrule
SigLIP \cite{chen2023pali}      & 	1,705    & I  & 58.9 / 57.8  & 68.7 / 68.5  & 2.58 / 2.26    & 37.2 / 42.5   & 1.14 / 1.02  & 49.6 / 52.6 \\ 
DinoV2-L \cite{oquab2023dinov2}    & 303  & I  & 62.3 / 62.0 & 64.6 / 67.6     & 1.96 / 2.05          & 53.9 / 56.0           & 0.77 / 0.74 & 56.0 / 60.3 \\
DinoV2-g \cite{oquab2023dinov2}    & 1,135 & I  & 63.9 / 62.3 &  67.0 / 68.8 & 2.00 / 2.06 & 47.5 / 50.4            & 0.67 / 0.66 & 55.5 / 59.8 \\
\midrule
VideoPrism-B \cite{zhao2024videoprism} & 114  & V  & 64.4 / 64.3  & 61.3 / 63.9         & 0.70 / 0.68          & 76.1 / 77.4 & 0.99 / 0.94           & 71.3 / 74.2 \\ 
VideoPrism-g \cite{zhao2024videoprism} & 1,113  & V  & 68.9 / 68.3  & 71.7 / \textbf{72.7}       & 0.75 / 0.73          & 77.6 / 78.5 & 0.81 / 0.78  & 71.8 / 74.3 \\
V-JEPA-L \cite{bardes2023vjepa}    & 307  & V  & 71.5 / 71.1  & 60.3 / 62.6        & 0.28 / 0.26        & 80.7 / 82.4            & 0.87 / 0.82           & 78.3 / 79.9 \\ 
V-JEPA-H \cite{bardes2023vjepa}    & 635  & V  & \textbf{73.4} / 72.7 & 61.5 / 63.7 & 0.28 / 0.27      & 78.0 / 81.4           & 0.79 / 0.76          & 78.8 / 80.9 \\ 
VideoMAE-B \cite{tong2022videomae}   & 87 & V  & 57.2 / 58.0   & 41.2 / 45.0 & 0.33 / 0.30           & 82.1 / 83.6 & 1.34 / 1.24          & 76.4 / 78.2 \\ 
VideoMAE-L \cite{tong2022videomae}  & 305  & V  & 66.7 / 66.8    & 53.8 / 57.0       & 0.27 / 0.28 & 81.5 / 82.3 & 0.91 / 0.85     & 78.7 / 80.9 \\
VideoMAE-H \cite{tong2022videomae}   & 633  & V  & 67.4 / 67.5   & 56.8 / 58.2         & 0.30 / 0.29   & 79.7 / 81.1            & 0.85 / 0.79 & 77.5 / 79.8 \\ 
VideoMAEv2-g \cite{wang2023videomae}~~~ & 1,013   & V & 70.5 / 69.9 & 70.5 / 70.3 & 0.39 / 0.38         & 80.4 / 82.0      & 0.84 / 0.79 & 77.0 / 80.0 \\ 
\midrule
\midrule
\Model-S & 24  & V &  43.9 / 46.0  & 30.8 / 33.9     & 0.59 / 0.51 & 77.4 / 78.3 & 1.89 / 1.72           & 73.5 / 75.7  \\
\Model-B & 91  & V & 54.9 / 55.4  & 38.6 / 42.8    & 0.43 / 0.41  & 81.7 / 82.9 & 1.38 / 1.30           & 76.3 / 78.5  \\
\Model-L & 310  & V &  63.8 / 64.3 & 48.7 / 52.1  & 0.33 / 0.30 & 83.9 / 85.2 &   0.99 / 0.93     & 79.2 / 81.1    \\
\Model-H & 639  & V & 66.4 / 66.6  & 51.9 / 54.4     & 0.31 / 0.29 & 84.6 / 85.6 & 0.89 / 0.83         & 80.1 / 82.0    \\
\Model-G & 1,848  & V & 68.5 / 68.5  & 55.3 / 57.8     & 0.25 / \textbf{0.24} & 85.1 / 85.9 & 0.75 / 0.70           & 81.0 / 82.4   \\
\Model-e & 3,811  & V & 69.6 / 69.0  & 55.2 / 60.4     & \textbf{0.24} / \textbf{0.24} & 85.1 / 85.8 & 0.68 / 0.65 & 81.5 / 82.8  \\
\Model-j & 21,495 & V & 72.5 / 72.0 & 59.2 / 62.9 & 0.25 / 0.26 & 86.3 / \textbf{86.6}   &  0.59 / \textbf{0.57}  & 81.3 / \textbf{83.0} \\
\bottomrule
\end{tabular}
\label{tab:baselines_longer_ft}
\end{table*}

\end{document}